\documentclass[conference]{IEEEtran}

\usepackage{amsmath,amsfonts,bm}

\def\eqref#1{equation~\ref{#1}}

\def\1{\bm{1}}

\DeclareMathAlphabet{\mathsfit}{\encodingdefault}{\sfdefault}{m}{sl}
\SetMathAlphabet{\mathsfit}{bold}{\encodingdefault}{\sfdefault}{bx}{n}

\usepackage{hyperref}
\usepackage{url}
\usepackage{tabularx}
\usepackage{multirow}
\usepackage{float}
\usepackage{subcaption}
\usepackage{tikz}
\usepackage[normalem]{ulem}
\usepackage{amsmath}
\usepackage{pifont}

\newcommand{\minitab}[2][l]{\begin{tabular}{#1}#2\end{tabular}}
\newcommand{\subhead}[1]{\vspace{3pt} \noindent {\bf #1}}
\newcommand{\secspace}{\vspace{0pt}}
\newcommand{\subsecspace}{\vspace{0pt}}
\newcommand*\circled[1]{\tikz[baseline=(char.base)]{\node[shape=circle,fill,inner sep=1pt] (char) {\textcolor{white}{#1}};}}
\newcommand{\etal}{et al.\ }
\newcommand{\pv}{{PerfVec }}
\newcommand{\pve}{{PerfVec}}
\newcommand{\uarch}{{microarchitecture }}
\newcommand{\uarchs}{{microarchitectures }}
\newcommand{\Uarch}{{Microarchitecture }}
\newcommand{\uarche}{{microarchitecture}}
\newcommand{\uarchse}{{microarchitectures}}
\newcommand{\Uarche}{{Microarchitecture}}
\newcommand{\Uarchse}{{Microarchitectures}}
\newcommand{\rep}{{representation }}
\newcommand{\reps}{{representations }}
\newcommand{\repe}{{representation}}
\newcommand{\repse}{{representations}}

\newcommand{\Reps}{{Representations }}
\newcommand{\gbl}{{generalizable }}
\newcommand{\cmark}{\ding{51}}%
\newcommand{\xmark}{\ding{55}}%

\begin{document}

\title{Learning Generalizable Program and Architecture \Reps for Performance Modeling}


\author{Lingda Li, Thomas Flynn, Adolfy Hoisie\\
Brookhaven National Laboratory\\
Upton, NY, USA \\
\texttt{\{lli, tflynn, ahoisie\}@bnl.gov} \\
}

\maketitle

\begin{abstract}
Performance modeling is an essential tool in many areas, including performance
characterization/optimization, design space exploration, and resource
allocation problems, to name a few.
However, existing performance modeling approaches have limitations, such as
high computational cost for discrete-event simulators, narrow flexibility of
hardware emulators, or restricted accuracy/generality of analytical/data-driven
models.
%
To address these limitations, this paper proposes \pve, a novel deep
learning-based performance modeling framework that learns high-dimensional and
independent/orthogonal program and \uarch \repse.
Once learned, a program \rep can be used to predict its performance on any
\uarche, and likewise, a \uarch \rep can be applied in the performance
prediction of any program.
Additionally, \pv yields a foundation model that captures the performance
essence of instructions, which can be directly used by developers in numerous
performance modeling related tasks without incurring its training cost.
The evaluation demonstrates that \pv is more general and efficient than
previous approaches.
\end{abstract}

%



\secspace
\section{Introduction}
\label{sect:intro}
\secspace

Performance modeling has been a practical area of emphasis and a computer
science and engineering challenge since the beginning of the computing era.
First, it is essential in performance characterization, verification, and
optimization.
It also provides a quantitative tool for architecture design space exploration,
as well as codesign of architectures, applications, and system software to
metrics of interest such as
performance, power, or resilience.
In addition, performance modeling is crucial to solve resource allocation and
scheduling problems, e.g., to improve the utilization of high performance
computing (HPC) systems by finding the most efficient hardware for a workload.

Performance modeling approaches can be broadly categorized into four classes.
The first involves simulating the program execution using discrete-event
techniques and software simulators \cite{gem5, sst}.
Simulation is the default means in early hardware design stages due to its
flexibility and ease of use, but its speed is plagued by high computational
cost and limited parallelism.
It is impractical to simulate real-world programs as a result.
Second, performance can be evaluated using hardware emulators whose speed is
close to native hardware \cite{MICRO07:Chiou, ISCA18:FireSim}.
However, emulators are expensive to build and lack flexibility, because
hardware implementation and verification require significant human expertise
and efforts.
The third and fourth categories are analytical and data-driven modeling.
Compared to simulators and emulators, analytical models are much faster because
they use high-level abstractions, which are based on simplified assumptions of
programs and \uarchse.
Analytical models try to capture various performance impact factors in abstract
equations \cite{CACM09:Roofline, TOCS09:Eyerman, TC16:Steen, Computer09:Barker,
IJHPCA00:Hoisie}.
Due to the human-engineered nature of analytical models and the complexity of
modern computers, they often ignore critical execution details and thus fall
short in accuracy.
Alternatively, recent work constructs data-driven models using machine learning
(ML) \cite{ASPLOS06:Ipek, ASPLOS06:Lee, MICRO15:Ardalani, HPCA15:Wu, DAC16:Li,
TIST14:Chen, MICRO07:Dubach}.
The most important drawback of existing ML-based models is they are program
and/or \uarch specific.
This means that when target programs or \uarchs change, new models must be
trained, and consequently, new datasets must be acquired for training.
Both data acquisition and model training are extremely time consuming and
computationally expensive, limiting their practicality.

In this work, we propose {\em \pve}, a data-driven performance modeling
framework that is \gbl to any program and \uarche.
\pv autonomously separates the performance impact of programs and \uarchs by
learning {\em program and \uarch \repse} that are {\em independent} of and {\em
orthogonal} to each other.
In ML terminology, \repse /embeddings are learned high-dimensional vectors that
represent certain types of data (e.g., words \cite{NIPS13:W2V},
images \cite{TPAMI17:SegNet}, proteins \cite{AlphaFold}) for general
understanding or specific tasks (e.g., language translation, image
segmentation) \cite{TPAMI13:RL}.
Representation learning has been one of the core ideas behind ML's great
success.
\pv employs separate models that aim to extract the key
and essential performance characteristics from programs and \uarchse,
respectively.

\pv has four key advantages.
First, \pv is the first {\em program and \uarch independent/oblivious} ML-based
performance model, while existing ML-based modeling approaches need to create
separate models for different programs or \uarchse.
In \pve, once a program's \rep is learned, it can be used to predict the
performance of that program on any \uarche.
Conversely, once a \uarche's \rep is learned, it can be used to predict the
performance of any program running on that \uarche.
Second, high-dimensional \reps produced by \pv can {\em capture subtle
performance-relevant details} compared with traditional human-engineered
analytical models, thereby resulting in better prediction accuracy.
Third, training \pv yields an instruction \rep model that captures
the essential performance characteristics of instructions.
Because this model can be used to learn the \uarch independent \rep of any
program compiled to an instruction set architecture (ISA), we believe it can
serve as a {\bf foundation} model for broad performance analysis/optimization
tasks, including but not limited to those demonstrated in Section
\ref{sect:case}.
Users can directly apply the pre-trained \pv foundation model in their tasks
without having to pay for its training overhead, in the same way that large
language models (LLMs) are applied across neural language processing (NLP)
tasks.
Last but not least, learning a program \rep is done by learning those of its
executed instructions in parallel.
Accelerators such as GPUs and HPC systems are well positioned to exploit such
{\em massive parallelism} to accelerate \pve.
\pv is publicly available at \url{https://github.com/PerfVec/PerfVec}.

The contributions detailed in this paper are:
\begin{itemize}
\item We devise a novel performance modeling framework that autonomously learns
to separate the performance impact of programs and \uarchse. (Section
\ref{sect:moti})
\item We create a foundation model to learn an instruction's \rep using
neighboring instructions that it interacts with during execution.
We rigorously prove that the \rep of a program can be composed by the \reps of
all its executed instructions.
The compositional property allows the foundation model to be generalizable to
any program. (Section \ref{sect:rep})
\item
To efficiently train the foundation model, we propose \uarch sampling and \rep
reuse.
Together, they reduce the training time by orders of magnitude. (Section
\ref{sect:train})
\item
The evaluation shows that \pv can make accurate performance predictions for
programs and \uarchs that are not present during training, demonstrating its
generality.
(Section \ref{sect:eval})
\item
We demonstrate two applications of \pv including design space exploration and
program analysis.
\pv significantly accelerates design space exploration compared to existing
approaches. (Section \ref{sect:case})
\end{itemize}

\secspace
\section{Toward Generalizable Performance Modeling}
\label{sect:moti}
\secspace

\begin{figure}[t]
\centering
\includegraphics[width=\linewidth]{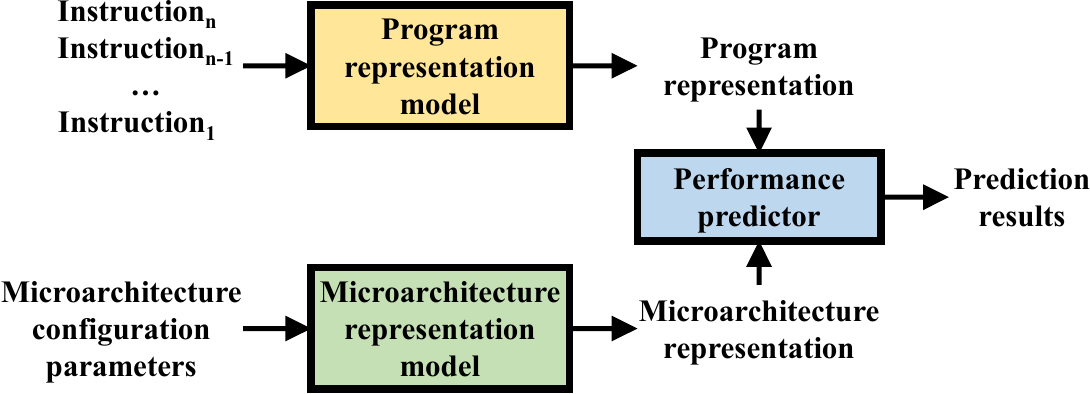}
\caption{The proposed \pv framework based on independent and orthogonal program
and \uarch \repse.}
\label{fig:arch}
\end{figure}

Performance is determined by the running program, the underlying \uarche, and
their interplay.
As introduced in Section \ref{sect:intro}, a performance model captures these
factors either in a complete simulation/emulation run of all instructions using
a detailed \uarch model, by a simplified mathematical model in human-engineered
analytical modeling, or through an ML-based surrogate in data-driven modeling.
Each of these methodologies has its advantages and limitations.
Simulation/emulation requires significant running time/human efforts, and
human-engineered analytical models have low prediction accuracy as they fail to
capture subtle execution details.
While ML-based models are much faster than simulation/emulation and also can be
more accurate than human-engineered analytical models, they have poor
generality as a trained model only works for certain programs and/or \uarchse.
Training a new model when target programs and/or \uarchs change is costly
because acquiring new training data is time-consuming, and training itself is
also computationally expensive.

This paper aims to solve the generality problem of existing ML-based models.
A \gbl performance model should have a {\em clear separation} between the
impact of programs and \uarchse.
Such ``separation of concerns'' minimizes the changes needed for the model when
target programs or \uarchs alter, and thus make it generalizable.
To achieve this goal, we devise {\em \pv} performance modeling framework that
autonomously separates the performance impact of programs and \uarchse.
Figure \ref{fig:arch} shows the overall architecture of \pve.
It contains three ML models.
First, a program \rep model learns \uarche-independent program performance \reps
from its execution trace.
The desirable \rep of a program should be able to predict its performance on
any \uarche.
Note that the program \rep depends on both the static program code and its
input.
Meanwhile, a counterpart \uarch \rep model learns program-independent \uarch
\reps from \uarch parameters.
Similarly, \uarch \reps should be applicable to any program.
Finally, a performance predictor makes predictions based on program and \uarch
\reps to model their interplay.

\secspace
\section{Learning Program Representations}
\label{sect:rep}
\secspace

\subsecspace
\subsection{Challenges}
\label{sect:rep:challenge}
\subsecspace

The biggest challenge of constructing \pv is building the 
program \rep model.
Previous ML-based models make use of high-level static (e.g., data/control flow
graphs \cite{ICS23:Trumper}) and/or runtime (e.g., performance counters
\cite{HPCA15:Wu, ICS23:Trumper}) information to learn \reps for performance
prediction.
Because of the complexity of modern computing systems, it is impossible to find
a complete set of high-level information that includes all performance relevant
execution details (e.g., memory level parallelism).
As a result, using such information as model input likely sacrifices prediction
accuracy.

While it is impossible to accurately derive the performance of a program from
its high-level properties, down to the fundamental machine instruction level,
the performance of any program comprehensively depends on by all its executed
instructions.
Therefore, it is natural to use the instruction execution trace to learn
accurate program \repse.

However, there is a reason why previous ML-based performance models do not work
on instruction execution traces,
because programs typically execute at least billions of instructions.
Contemporary ML models struggle when processing long input sequences due to the
vanishing gradient problem in recurrent neural networks (RNNs)
\cite{TNN94:Bengio} and the quadratic attention computational cost of
Transformers \cite{NeurIPS22:FlashAttention}.
For example, state-of-the-art LLMs typically take thousands of input tokens
(e.g., up to 32k for GPT-4 \cite{GPT4token}), which is nowhere close to more
than billions of tokens in the performance modeling scenario being investigated
herein.
Existing ML-based performance models that use instruction execution traces can
only predict the performance of basic blocks with a handful of instructions due
to this restriction \cite{ICML19:ithemal, IISWC22:GRANITE}.

In this work, we will demonstrate \pv can learn the \rep of a program using its
complete execution trace with no limit on the amount of instructions it
contains, through the careful design of {\em compositional} \repse.

\subsecspace
\subsection{Compositional Instruction Representations}
\label{sect:rep:method}
\subsecspace

\begin{figure}[t]
\centering
\includegraphics[width=\linewidth]{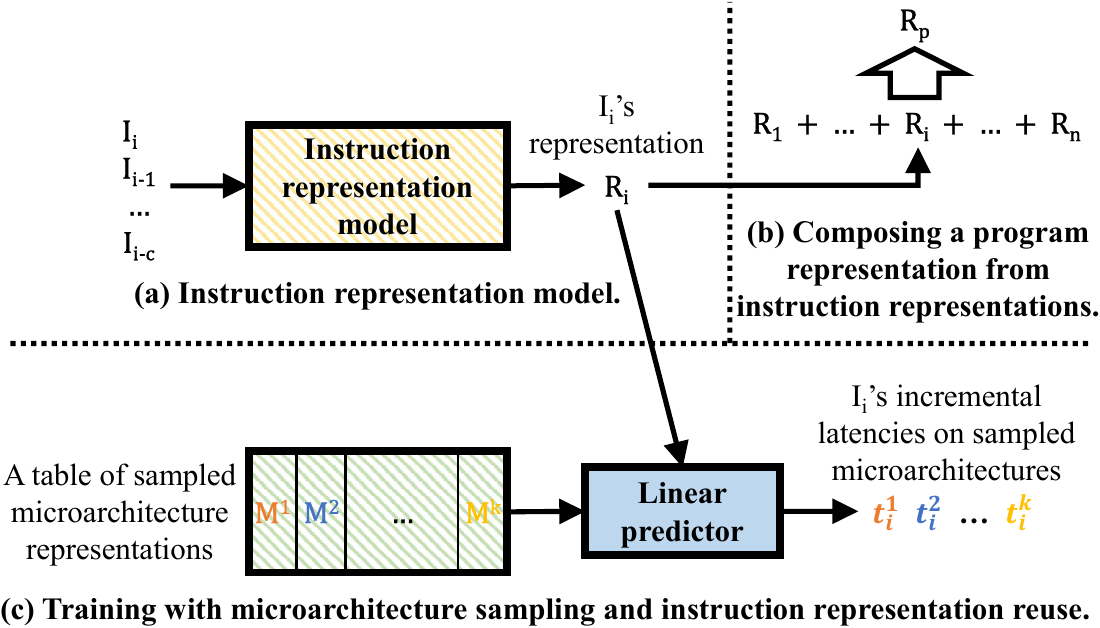}
\caption{The framework to learn instruction \repse.}
\label{fig:instrep}
\end{figure}


While it is infeasible to directly learn program \reps from billions/trillions
of instructions, it is within the capability of today's ML models to predict
the performance of a single instruction and learn its \repe.
Despite the extreme complexity of modern processors, the execution and
performance of an instruction is mostly affected by hundreds of instructions
issued within a relatively narrow window before it (``neighboring''
instructions) due to the instruction level parallelism constraint
\cite{ASPLOS91:ILP}.
Capitalizing on this fact, we can construct ML models to predict the execution
latency of an instruction using the properties of itself and neighbors.
Note that the performance of an instruction also can be affected by older
instructions through caches, etc.
Section \ref{sect:rep:feature} will describe how we capture these impacts
through input features.

Figure \ref{fig:instrep} depicts the framework to learn instruction \repse.
Compared with Figure \ref{fig:arch}, it replaces the program \rep model with
the instruction \rep model.
This model takes the current instruction $I_i$ and previous $c$ instructions as
inputs and outputs $I_i$'s \rep $\bm{R}_i$.
Sections \ref{sect:rep:feature} and \ref{sect:rep:model} will introduce more
details about the instruction \rep model.

The key idea of \pv is to compose a program \rep from the \reps of its executed
instructions.
We carefully design the \pv framework and its prediction targets for this
purpose.
The first key choice is for the framework to predict the {\em incremental
latency} of an instruction, 
which is defined as the time length that an instruction stays active in the
processor after all of its predecessors exit.
For example, if instructions $I_1$, $I_2$, and $I_3$ retire at 50, 100, and 120
ns respectively, the incremental latency of $I_3$ is $120 - 100 = 20$ ns.
Note that the incremental latency of an instruction may be zero when it retires
earlier than its predecessors (this can happen for certain \uarchse) or at the
same time.

%

To enable the composition of program \reps from instruction \repse, the second
key choice involves selecting the performance predictor (the blue box in Figure
\ref{fig:instrep}(c)) to be a {\em linear} model without additive biases.
With these two primary choices, we now prove that the \rep of a program can be
obtained by summing the \reps of all its instructions as follows.

Assume a program executes $n$ instructions in total, and $t_i$ and $\bm{R}_i$
are the incremental latency and \rep of the $i$th instruction $I_i$.
Because the predictor is a linear model, we get $t_i = \bm{R}_i \ast \bm{M}$,
where $\bm{M}$ is the target \uarch \rep and $\ast$ is the dot product
operation.
According to the definition of incremental latency, the total execution time
$T$ of this program can be calculated by accumulating all incremental
latencies:
\begin{align*}
T &= \sum_{i=1}^{n} t_i = \sum_{i=1}^{n} (\bm{R}_i \ast \bm{M}) ={\circled{1}} \sum_{i=1}^{n} (\sum_{j=1}^{d} (r_{ij} \times m_j)) \\
&={\circled{2}} \sum_{j=1}^{d} (\sum_{i=1}^{n} (r_{ij} \times m_j)) ={\circled{3}} \sum_{j=1}^{d} ((\sum_{i=1}^{n} r_{ij}) \times m_j) \\
&={\circled{4}} (\sum_{i=1}^{n} \bm{R}_i) \ast \bm{M},
\end{align*}
where $d$ is the \rep dimensionality.
\circled{1} and \circled{4} employ the definition of dot product,
\circled{2} uses the commutativity of sums, and \circled{3} extracts common
factors $m_j$.

Because $T = (\sum_{i=1}^{n} \bm{R}_i) \ast \bm{M}$, we may interpret the sum $\sum_{i=1}^{n} \bm{R}_i$
as the \rep of this program $\bm{R}_p$.
That is, the program \rep is equal to the summation of \reps of all its
executed instructions, as shown in Figure \ref{fig:instrep}(b), and a program's
total execution time is predicted using the dot product of the program and
\uarch \repse.
Notably, this compositional property only holds when the performance predictor
is a linear model and the training target is integrable (i.e., the total
execution time can be calculated by summing incremental latencies).

One may wonder whether the use of a fixed linear predictor will affect the
predictive ability of \pve.
Fortunately, \pv is still capable of capturing the nonlinearity of performance
prediction through deep learning-based instruction and \uarch \rep models.
Our choice of using a linear performance predictor is also justified by
previous human-engineered analytical models \cite{TOCS09:Eyerman, TC16:Steen,
SC02:Snavely}.
Their manually designed equations to calculate execution time share similar
forms of dot products of two vectors.

\begin{table*}[t]
\footnotesize
\centering
\caption{Input features for the instruction \rep model.}
\label{tbl:features}
\begin{tabularx}{\linewidth}{|l|X|}
\hline
{\bf Category} & {\bf Features} \\
\hline
\hline
Static property & 15 operation features (operation type, direct/indirect branch, memory barrier, etc.); indices and categories for 8 source and 6 destination registers \\
\hline
Execution behavior & Execution fault or not (e.g., divide by zero); branch taken or not \\
\hline
Memory access & Stack distances for instruction fetch; stack distances with respect to all data accesses, all loads, and all stores \\
\hline
Branch predictability & Global branch entropy that considers the taken/untaken history of all branches; local branch entropy that considers the history of the current branch \\
\hline
\end{tabularx}
\end{table*}


\subhead{Advantages of Compositional Representations.}
Our compositional approach that constructs program \reps from instruction \reps
has several key benefits.
First and foremost, as all programs compiled to the same ISA can be viewed as
different combinations of the same set of instructions, we only need to train
the instruction \rep model {\em once}.
Then, it is {\em \gbl to any program} to learn its \repe.
We call this instruction \rep model the {\bf foundation} model of \pv because
of its wide applicability, as partially demonstrated in Section
\ref{sect:case}.
Second, compositional \reps enable not only overall but also {\em detailed
analysis}.
Moreover, the \reps of all instructions can be generated in parallel when
calculating the \rep of a program, which yields a massive amount of
parallelism.
HPC systems equipped with accelerators such as GPUs and TPUs can exploit this
parallelism to reduce the time needed to construct a program's \repe.

\subsecspace
\subsection{Instruction Features}
\label{sect:rep:feature}
\subsecspace

The instruction \rep model requires proper input features to learn the
performance characteristics of instructions.
These features should be sufficient to capture all performance impacts, and
they should also be {\em \uarche-independent} because we would like learned
instruction \reps to be \gbl across \uarchse.

Table \ref{tbl:features} lists all 51 instruction features.
Static instruction properties, such as operation types and register indices,
are included, as well as \uarche-independent dynamic execution behaviors, e.g.,
whether or not a branch instruction jumps to its target.

The performance of an instruction is also affected by components that have
long-lasting states, including caches, translation lookaside buffers, and
branch predictors.
Traditionally, people use \uarche-dependent features (e.g., cache miss number)
to model these effects \cite{TOCS09:Eyerman, SIGMETRICS22:simnet}, which are
not desirable here because the goal is to learn \uarche-independent \repse.

Fortunately, several \uarche-independent properties have been proposed for
memory access and branch prediction analysis, which, in turn, can be used as
the input features for the instruction \rep model.
To model the impact of memory accesses, we use {\em stack distance}
\cite{PLDI03:Ding, HPCA16:Beckmann, IISWC13:Byfl}.
The stack distance of an access is defined as the number of unique memory
accesses between the current and last accesses to the same address.
Intuitively, accesses with longer stack distances are more likely to miss in
caches.

To model the branch predictability, we employ {\em branch entropy}
\cite{ARCS08:BE, TC17:LBE}.
In this method, we use 1 to denote a branch taken and 0 for a branch untaken.
Thus, the branch taken history can be represented as a sequence of 0s
and 1s.
Then, an entropy is calculated on this number sequence, called branch entropy.
Branches that show more consistent patterns (e.g., always taken or always not
taken) have lower entropies and are easier to predict.
For example, a branch that always jumps to the target has only 1s in the
number sequence.
As a result, its branch entropy is 0.
The entropies of a branch are calculated both locally (i.e., only use the
history of the current branch) and globally (i.e., include the history of other
recent branches) for input features.

\subsecspace
\subsection{Model Architecture}
\label{sect:rep:model}
\subsecspace


As illustrated in Figure \ref{fig:instrep}(a), the input of the foundation
model is a sequence of $c+1$ instructions.
$c$ determines the number of instructions to look back.
Instructions that are closer to the current instruction are likely to have
larger impact on its performance, while the impact descends for faraway
instructions.
In practice, we find that having $c=255$ is enough to capture most impacts of
neighboring instructions and considering more instructions is unnecessary.
Given that each instruction has 51 features, there are $51 \times (255+1) =
13056$ input features in total.

Regarding ML model architectures, many have been invented to address sequences.
Section \ref{sect:eval:ab} will compare various sequence processing models.
Our exploration finds a 2-layer 256-dimensional unidirectional long short-term
memory (LSTM) model performs well enough, and more complex models do not bring
significant benefits.
Notably, \pv users treat the pre-trained foundation model as a black box and
its architecture is irrelevant to them as will be shown in Section
\ref{sect:case}.

%

\secspace
\section{Training the Foundation Model}
\label{sect:train}
\secspace

\subsecspace
\subsection{\Uarch Sampling}
\label{sect:train:sample}
\subsecspace

There are several computational challenges in training the instruction \rep
model presented in Section \ref{sect:rep}.
As suggested by Figure \ref{fig:arch}, to obtain an instruction \rep model, a
\uarch \rep model needs to be jointly trained.
However, training a \uarch \rep model is not trivial because of the huge \uarch
design space.
Describing a conventional \uarch requires thousands of parameters, and thus the
\uarch design space has at least $10^{1000}$ configurations.
Training a reliable model to represent this immense design space poses a
significant challenge.
In addition, it is also unnecessary to do so as computer architects constantly
introduce new \uarchse.
Hence, even if we were able to train a complete \uarch \rep model, it will
likely be outdated soon.

%

In this work, we are able to train the instruction \rep model without having to
train a \uarch \rep model jointly through {\em \uarch sampling}.
As shown on the left side of Figure \ref{fig:instrep}(c), we sample $k$ \uarchs
whose to-be-learned representations $\bm{M}^1, ..., \bm{M}^k$ are stored in a
table.
The instruction \rep model is trained to predict $I_i$'s incremental latency
$t_i^j$ on the $j$th \uarche, and $\bm{M}^j$ is also trained along.
This process repeats for the combinations of all instructions and $k$ sampled
\uarchse.

\Uarch sampling allows us to train the \reps of sampled \uarchs $\bm{M}^1, ...,
\bm{M}^k$ instead of a \uarch \rep model that generates \reps from
configuration parameters.
The hypothesis is that with sufficient, diverse, and representative \uarch
samples, the instruction \rep model should be able to learn \uarche-independent
performance characteristics and generalize to unseen \uarchse.
Section \ref{sect:eval:general} will evaluate this hypothesis.

Training with \uarch sampling significantly lowers the computational cost and
speeds up the convergence because there are fewer parameters to train.
As will be introduced soon, we sample 77 \uarchs in total, resulting in $77
\times 256 = 19.7k$ \uarche-related trainable parameters, assuming
256-dimensional \repse.
Alternatively, assuming to train a basic hypothetical 2-layer neural network
\uarch \rep model with 1000 input parameters and 1000 hidden neurons, there are
$1000 \times 1000 + 1000 \times 256 = 1.3$ million parameters to train, which
is $\sim60 \times$ bigger.
A credible \uarch \rep model likely needs to be orders of magnitude larger than
this simple model.

\subsecspace
\subsection{Instruction Representation Reuse}
\label{sect:train:reuse}
\subsecspace

While \uarch sampling eliminates the need of a \uarch \rep model, the training
overhead is still extremely high in the naive approach where the model predicts
the latency of an instruction on one \uarch at a time.
In this approach, the training time increases linearly with the number of both
instructions and sampled \uarchs in the training dataset.
Given the training dataset with 737 million instructions and 77 \uarchs that we
collect (see Section \ref{sect:train:data}), one epoch takes roughly 26 days to
go through all 58 billion combinations to train the default 2-layer LSTM using
an NVIDIA A100 GPU.
This is prohibitively slow even with multiple GPUs.

Because the instruction \rep model is the most computationally intensive part in
Figure \ref{fig:instrep}, we propose an efficient training procedure via the
{\em reuse of instruction \repse}.
The key fact is that for the same program and input, the logical instruction
execution trace does not change with the underlying \uarche.
Capitalizing on this fact, we execute the same program on all sampled \uarchs
to obtain instruction latencies of the same trace.
Using these results, given an instruction $I_i$, we let the model predict its
incremental latencies $t_i^1, ..., t_i^k$ on all $k$ \uarchs simultaneously, as
shown in Figure \ref{fig:instrep}(c).
During the forward pass of training, we compute $I_i$'s \rep $\bm{R}_i$ once
before combining it with $\bm{M}^1, ..., \bm{M}^k$ to calculate all $k$
latencies.
During backpropagation, the gradient decent is also performed only once on the
instruction \rep model by averaging gradients with respect to the prediction
errors of $t_i^1, ..., t_i^k$.
Through the reuse of $\bm{R}_i$, the expensive forward and backward passes of
the instruction \rep model are amortized.

Reusing instruction \reps cuts per-epoch training time from 26 days to 8 hours
on one GPU by reducing complexity from linear to near-constant with the number
of sampled \uarchse.

\subsecspace
\subsection{Dataset}
\label{sect:train:data}
\subsecspace

To collect a sufficient dataset to train the foundation model, we leverage gem5
\cite{gem5}, one of the most widely used simulators.
There are two practical reasons for choosing simulators instead of real
processors.
First, we need to obtain individual instruction latencies to train the model,
and real-world processors cannot provide such detailed information.
Second, it is more convenient for simulators to sample a diverse range of
\uarchse, while real-world processors have less diverse \uarchse.
In addition, it is not practical to purchase a large number of real CPUs and
build/maintain their evaluation environments.
Despite training with simulator data, \pv remains valuable for computer
architecture design due to the irreplaceable role of simulators in this field.
%

Upon the instruction execution trace produced by gem5, we implement a tool to
generate the desired input features listed in Table \ref{tbl:features} for
every instruction, 
and their incremental latencies (in the unit of 0.1 ns) as prediction targets.

\subhead{\Uarche.}
To ensure sufficient coverage of the \uarch space, we develop a tool to
randomly sample valid gem5 configurations.
This tool can alter processor, cache, and memory configurations.
For processor configurations, it can vary the processor type (i.e., in-order
vs. out-of-order), frequency, pipeline (e.g., fetch stage latency; issue
width), execution units (e.g., the number and latency of floating-point
multiplication units), and branch predictors.
For cache configurations, we can randomly select cache sizes, associativities,
latencies, and exclusivity.
It can also change the memory type (e.g., DDR4, LPDDR5, GDDR5, or HBM),
bandwidth, and frequency.
There are hundreds of tunable knobs.

Using this tool, we sample 60 out-of-order and 10 in-order configurations
randomly.
More out-of-order than in-order processors are sampled, as they are predominant
in today's computer systems and more challenging to model.
Together with seven predefined configurations in gem5 (four out-of-order and
three in-order), we collect a training dataset with 77 \uarchse.

\begin{table}[t]
\footnotesize
\centering
\caption{Benchmarks for ML training and testing.}
\label{tbl:benchmarks}
\begin{tabular}{|c|c|c|}
\hline
{\bf Type} & {\bf Training} & {\bf Testing} \\
\hline
\hline
INT & \minitab[c]{525.x264, 531.deepsjeng,\\ 548.exchange2, 557.xz,\\ 999.specrand} & \minitab[c]{500.perlbench, 502.gcc,\\ 505.mcf, 523.xalancbmk} \\
\hline
FP & \minitab[c]{527.cam4, 538.imagick,\\ 544.nab, 549.fotonik3d} & \minitab[c]{507.cactuBSSN, 508.namd,\\ 519.lbm, 521.wrf} \\
\hline
\end{tabular}
\end{table}

\subhead{Program.}
We adopt the widely used SPEC CPU2017 benchmark suite \cite{bucek2018spec} to
train and test \pv because of its representativeness.
These benchmarks are expected to cover a diverse range of instruction execution
scenarios, which should help train \gbl models.
In our environment, all benchmarks are compiled to ARMv8 under O3 of gcc 8.2
and 17 of them run successfully.
We divide these 17 benchmarks about evenly for training and testing (shown in
Table \ref{tbl:benchmarks}).
To ensure fairness, the division is decided based on the benchmark indices:
eight benchmarks with smaller indices are used for testing, and the other nine
benchmarks are used for training.
To collect the training dataset, each training benchmark is simulated using the
default test input by 100 million instructions on all 77 \uarchse.
Combining all simulation results, we obtain a dataset of 2.0 TB, among which
roughly 90\% of them are dedicated for training, 5\% for validation, and 5\%
for testing.

Note there are no restrictions on the amounts of programs and \uarchs used in
the training dataset, and a larger dataset improves generality and accuracy
(see Section \ref{sect:eval:ab}).

\subsecspace
\subsection{Training Setup}
\label{sect:train:setup}
\subsecspace

We implement \pve's training and testing framework using PyTorch.
The mean squared error between predicted and true latencies is used as the loss
function.
We adapt the Adam optimizer, where the initial learning rate is set to 0.001
and decayed by $10\times$ every 10 epochs.
Each model is trained for 50 epochs.
The validation set is used to choose the model with the lowest validation loss
among all epochs.

We train our models on several GPU servers.
With the training optimizations described in Sections \ref{sect:train:sample}
and \ref{sect:train:reuse}, it takes two days to train the default LSTM model
on eight A100 GPUs.
Notably, \pv users, like LLM users, are not burdened with this training
overhead because they can directly use the pre-trained foundation model or
pre-generated program \repse.

\secspace
\section{Evaluation}
\label{sect:eval}
\secspace

\subsecspace
\subsection{Accuracy and Generality}
\label{sect:eval:general}
\subsecspace

\subhead{Seen Programs and Seen \Uarchse.}
First, we evaluate the prediction accuracy of \pv for 9 ``seen'' programs and
77 ``seen'' \uarchs that appear in the training dataset.
As discussed in Section \ref{sect:rep:model}, the 2-layer 256-dimensional LSTM
model is adopted as the default foundation model in the following experiments.

\begin{figure}[t]
\centering
\includegraphics[width=\linewidth]{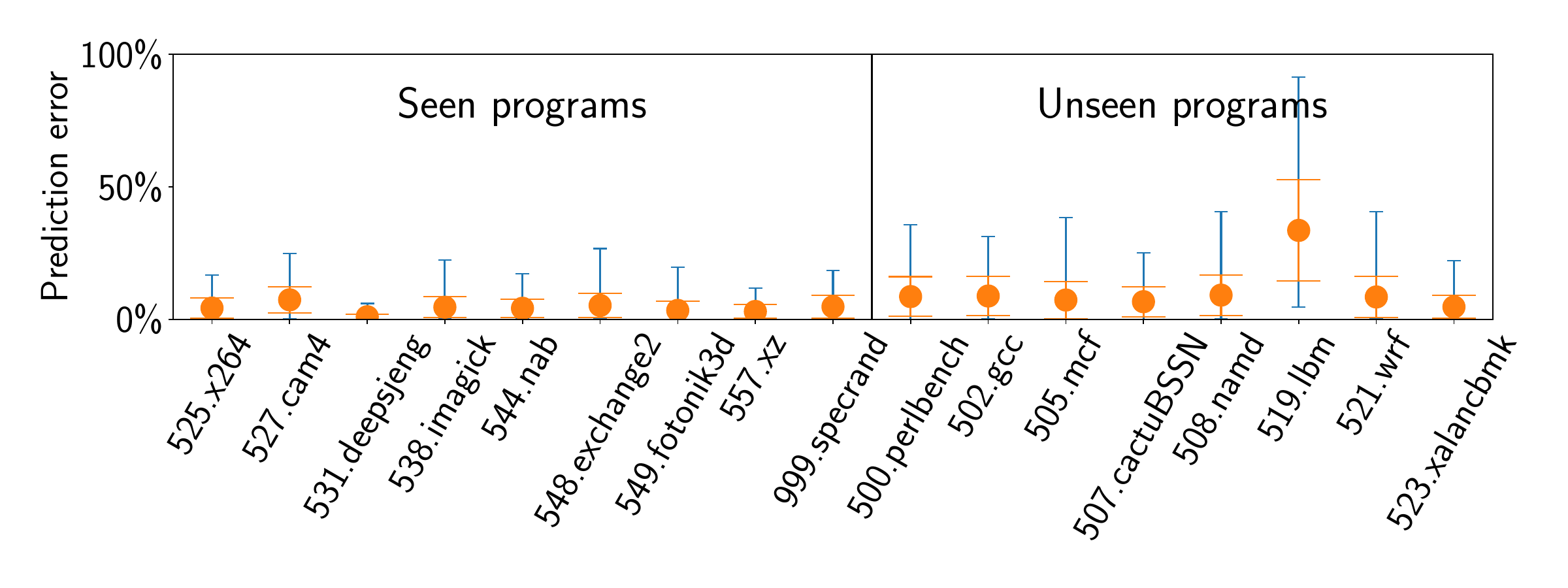}
\caption{Performance prediction accuracy for seen and unseen programs on seen \uarchse.}
\label{fig:seen_uarch}
\end{figure}

The left side of Figure \ref{fig:seen_uarch} shows the absolute errors on the
predicted execution time of \pv compared with gem5 simulation results per seen
program.
Dots denote the average absolute prediction errors across all 77 seen \uarchse,
orange caps mark standard derivations across \uarchse, and blue caps mark
maximum and minimum errors.
We observe the average errors are below 8\% for all programs, and the maximum
errors are also below 30\%, demonstrating \pve's accuracy on seen programs and
\uarchse.

\subhead{Unseen Programs and Seen \Uarchse.}
To achieve the desired generality, it is essential for \pv to be able to make
accurate predictions on ``unseen'' programs and \uarchs that are not available
during training.
First, we measure the prediction accuracy of \pv on unseen programs.
To generate the \rep of an unseen program, the \reps of all its executed
instructions are first generated by the pre-trained foundation model.
Then, summing them gives the program \rep as described in Section
\ref{sect:rep:method}.
Together with the \reps of 77 seen \uarchs learned from training, we can
predict the performance of unseen programs on seen \uarchse.

\begin{figure}[t]
\centering
\includegraphics[width=\linewidth]{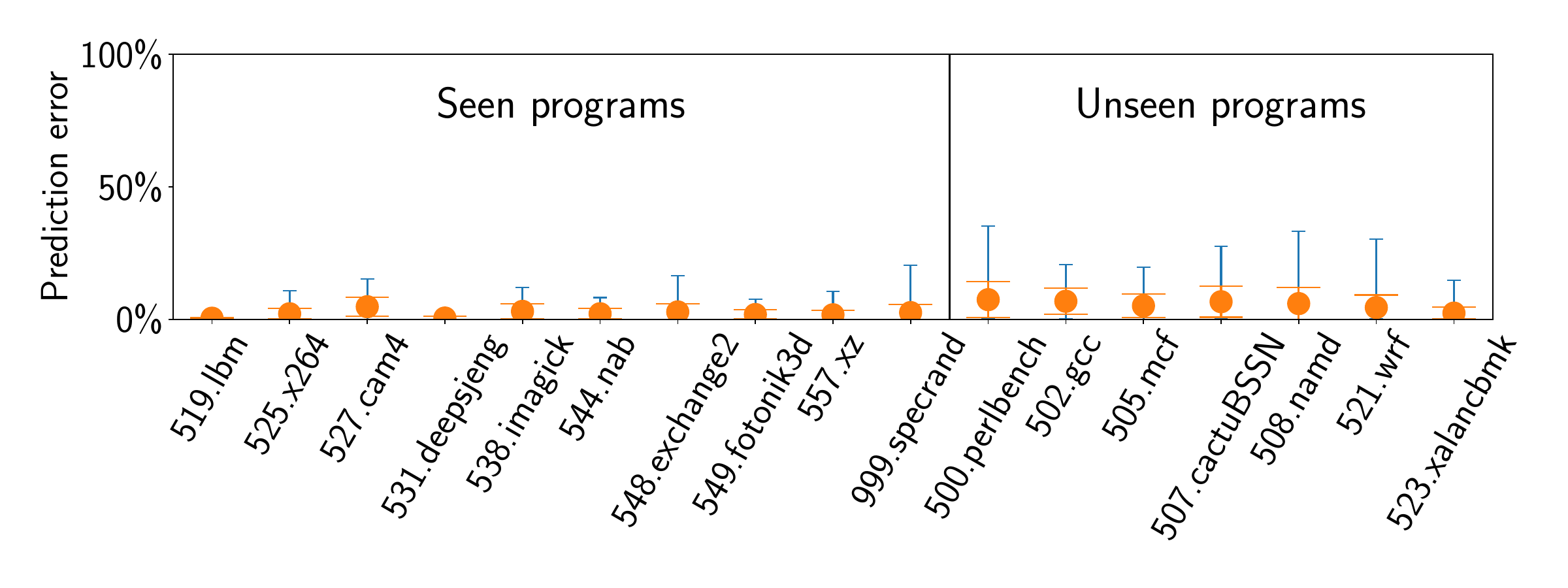}
\caption{Prediction accuracy on seen \uarchs after moving \texttt{519.lbm} to the training dataset.}
\label{fig:seen_uarch_lbm}
\end{figure}

The right side of Figure \ref{fig:seen_uarch} shows the prediction errors of
unseen programs.
Although the prediction errors increase compared with those of seen programs,
the average errors are below 10\% for most unseen programs with \texttt{519.lbm}
as an exception.
The main reason why \pv generalizes well to most unseen programs is that
program \reps are composed by instruction \repse, and all programs are
eventually composed by the same set of instructions, just different
combinations.
We hypothesize \texttt{519.lbm} incurs higher errors because the training
dataset lacks sufficient coverage of its instruction combination scenarios.
To test this hypothesis, we move \texttt{519.lbm} to the training dataset and
retrain the default model.
Figure \ref{fig:seen_uarch_lbm} shows the prediction errors of the model
trained on the updated dataset.
We observe that the errors of \texttt{519.lbm} effectively reduce close to zero.
Furthermore, the updated model improves the prediction accuracy of other seen
and unseen programs, compared with Figure \ref{fig:seen_uarch}.
This experiment shows that larger datasets have better instruction combination
coverage and thus improve the generality and accuracy of trained models, and
more evidences about this will be presented in Section \ref{sect:eval:ab}.
The updated model is used in the following experiments.


\begin{figure}[t]
\centering
\includegraphics[width=\linewidth]{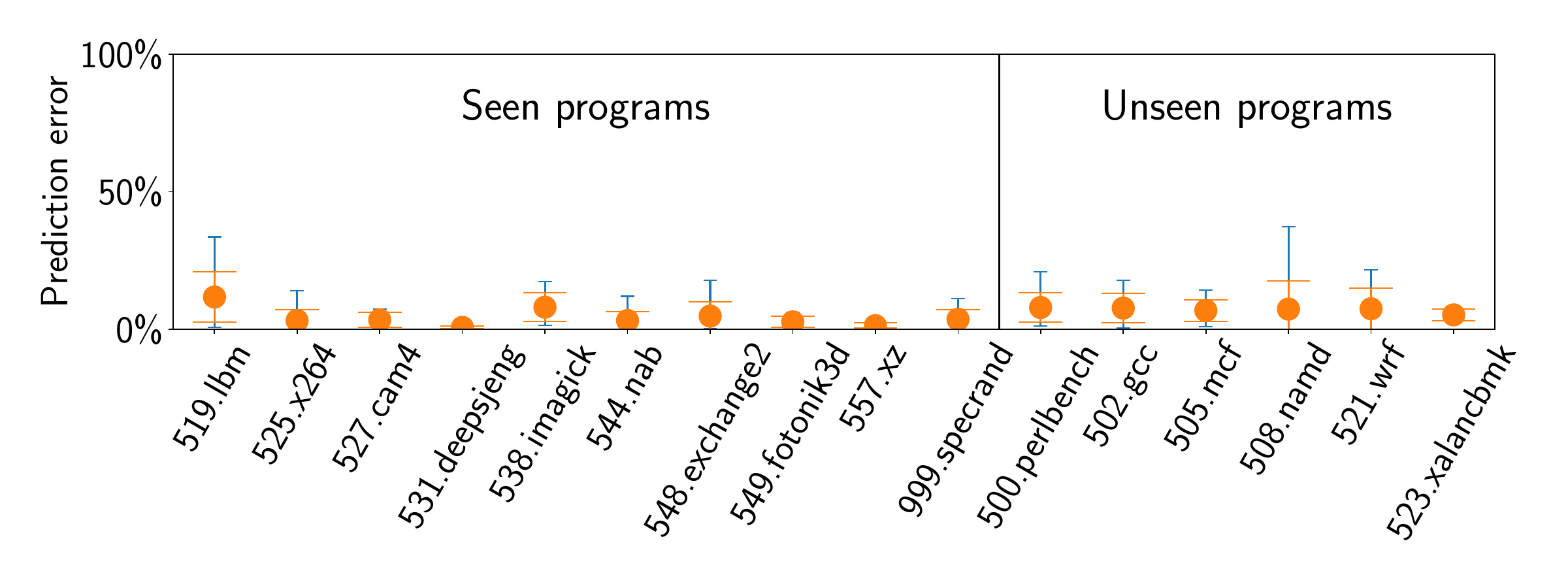}
\caption{Performance prediction accuracy for seen and unseen programs on unseen \uarchse.}
\label{fig:unseen_uarch}
\end{figure}

\subhead{Unseen \Uarchse.}
To test the generality of \pv to unseen \uarchse, we randomly generate 10 more
\uarch configurations that are not used in training and test if \pv can
accurately predict program performance on these \uarchse.
While it is trivial to obtain the \reps of both seen and unseen programs using
the pre-trained foundation model, we describe how to obtain the \reps of unseen
\uarchs for performance prediction as follows.

Initially, we obtain a small tuning dataset for the target unseen \uarchs by
simulating several seen programs.
Then, unseen \uarch \reps are learned using the framework shown in Figure
\ref{fig:instrep}, with an important difference that the instruction \rep model
is initilized to be the pre-trained foundation model and frozen during
training.
Only the \uarch \rep table is updated in backpropagation to learn the \reps of
unseen \uarchse.
This process can be viewed as a specific case of fine tuning, which has been
applied broadly to adapt general-purpose ML models for specific tasks
\cite{18:BERT}.


Figure \ref{fig:unseen_uarch} illustrates the prediction accuracy of all
programs on unseen \uarchse.
Because \texttt{507.cactuBSSN} incurs errors on an unseen \uarch in gem5
simulation, it is not included.
The average errors across seen and unseen programs are 4.2\% and 7.1\%,
respectively.
Individual programs' errors are comparable with those on seen \uarchs in Figure
\ref{fig:seen_uarch_lbm}.
Therefore, we conclude that \pv generalizes well to unseen \uarchs too.



\subsecspace
\subsection{Ablation Studies}
\label{sect:eval:ab}
\subsecspace

\begin{figure}[t]
\centering
\includegraphics[width=0.9\linewidth]{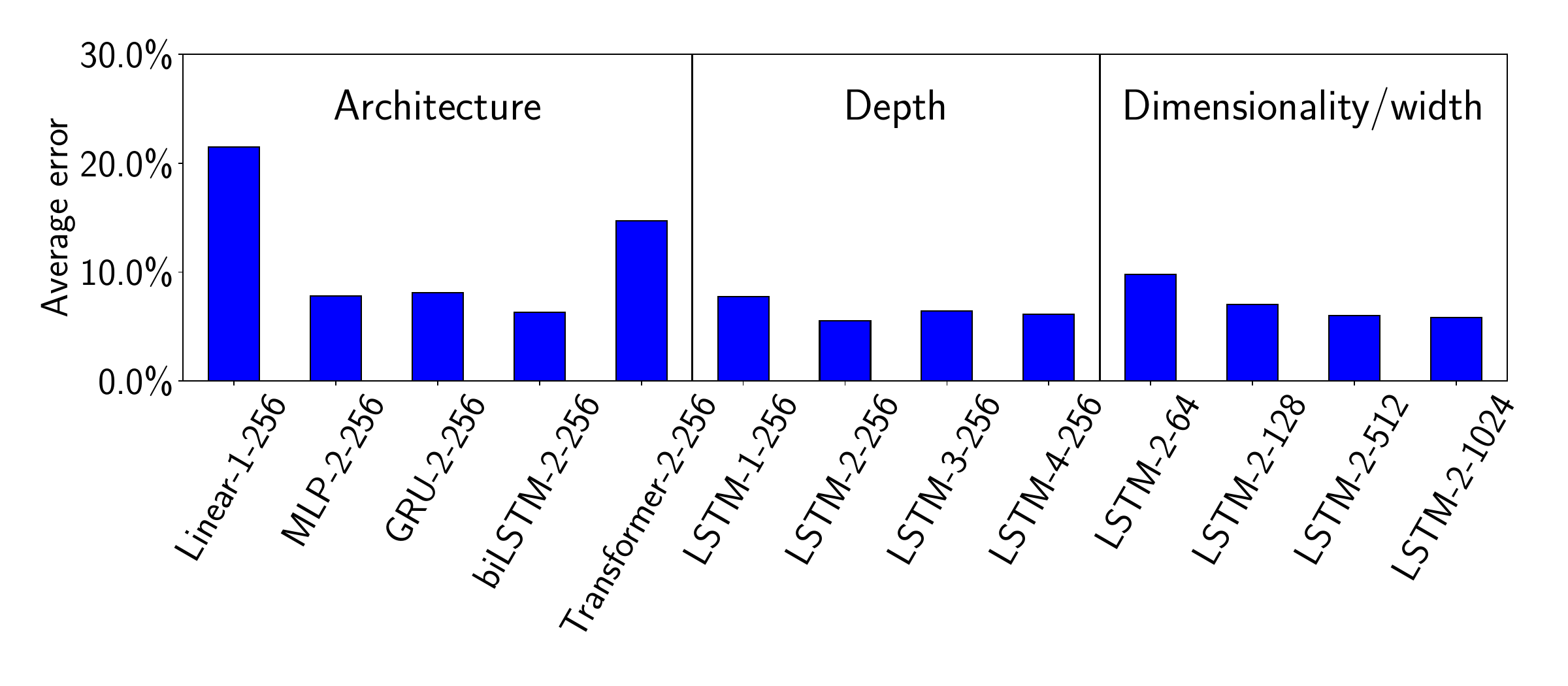}
\caption{Accuracy of various ML models.}
\label{fig:nas}
\end{figure}

\subhead{Model Architecture.}
Figure \ref{fig:nas} compares the prediction errors averaged across unseen
programs when altering the foundation model architecture.
The x axis shows various models, whose names are composed by three parts: the
model architecture, the number of layers, and the \rep dimensionality.
The leftmost portion of Figure \ref{fig:nas} explores neural architectures
other than the default unidirectional LSTM, including linear regression,
multilayer perceptron (MLP), gated recurrent unit (GRU), bidirectional LSTM
(biLSTM), and Transformer encoder.
The linear model incurs the largest error because it is insufficient to address
the complexity of performance modeling.
Transformer has the second largest error, and larger Transformer models fail to
converge in training.
Other models perform slightly worse than or similar to LSTM models shown on the
right.

The middle and right portions of Figure \ref{fig:nas} focus on LSTM and study
the impacts of the model depth and the \rep dimensionality, respectively.
Deeper and wider models reduce errors at the beginning, but using more than two
layers or higher than 256-dimensional \reps brings negligible or no accuracy
improvement.

These experiments suggest that the default 2-layer 256-dimensional LSTM model
(LSTM-2-256) is sufficient.
Note that our neural architecture search (NAS) is preliminary.
We believe that more thorough and systematic NAS will discover better \pv
models, which is left for future work.


\subhead{Training Data Volume.}
We also study the impact of training data volume by varying two factors: the
numbers of instructions and \uarch configurations it contains.
The average prediction error of the default LSTM model drops from 7.7\%, 5.2\%,
to 3.6\% as the number of instructions increases from 10\%, 50\%, to 100\% of
the default training dataset.
Decreasing the sampled \uarch number from 77 to 20 leads to a larger average
error increase on unseen \uarchs (from 5.3\% to 7.9\%) compared to that on
unseen programs (from 5.5\% to 7.2\%), confirming the claim in Section
\ref{sect:train:sample} that sampling more \uarchs improves the
generalizability of trained models to unseen \uarchse.
These experiments suggest that \pv models benefit from larger training data, in
terms of both instruction and \uarch amounts and diversities.
It is not surprising as similar phenomenon has been observed repeatedly in
other ML domains.
Thus a part of future work is to further enrich training data.

\subhead{\Uarche-Independent Features.}
The use of \uarche-independent instruction features is a crucial element that
enables \pve's generalizability across \uarchse, especially those related to
memory access and branch predictability.
To investigate their importance, we train and evaluate the default LSTM model
when memory and branch related features are removed from input.
Without these features, the average prediction error on unseen programs soars
to 17.0\%, compared to 5.5\% when using them.
This result highlights the critical role that these features play in \pv
because they are essential to capture memory and branch behaviors.

\subsecspace
\subsection{Comparison with State-of-the-Arts}
\label{sect:eval:compare}
\subsecspace

\begin{table*}[t]
\centering
\scriptsize
\caption{Comparison of various ML-based modeling and simulation approaches.
$^{\ast}$Training overheads are estimated qualitatively as some of them are
not reported.
$^{\dagger}$Prediction speed numbers are taken from original papers.
Note that different hardware platforms were used in these papers so this is not
a rigorous comparison.
\pve, program-specific, and transferable models can make instant program
performance predictions.
The speeds of other methods depend on the number of executed instructions and
are thus measured in terms of instruction per second (IPS).}
\label{tbl:comparison}
\begin{tabular}{|c|c|c|c|c|c|c|c|}
\hline
\multirow{2}{*}{Approach} & \multirow{2}{*}{Input} & \multirow{2}{*}{Target} & \multicolumn{2}{c|}{Overhead} & \multicolumn{2}{c|}{Generality} \\
\cline{4-7}
& & & Training$^{\ast}$ & Prediction$^{\dagger}$ & Program & $\mu$arch \\
\hline
Ithemal \cite{ICML19:ithemal}, GRANITE \cite{IISWC22:GRANITE} & Textual instruction trace & Basic block & Minutes & 500--50k IPS & \cmark & \cmark\kern-1.2ex\raisebox{.7ex}{\rotatebox[origin=c]{120}{\textbf{---}}} \\
\hline
Performance embedding \cite{ICS23:Trumper} & Control/data flow graph \& performance counters & Loop nest & Days & Billions IPS & \cmark & \xmark \\
\hline
Program-specific models \cite{ASPLOS06:Ipek, HPCA07:Lee, DAC16:Li} & N/A & Program & Days--weeks & $< 1$ s & \xmark & \xmark \\
\hline
Transferable models \cite{MICRO07:Dubach, PACT07:Khan, SC17:Marathe} & N/A & Program & Hours--days & $< 1$ s & \cmark\kern-1.2ex\raisebox{.7ex}{\rotatebox[origin=c]{120}{\textbf{---}}} & \xmark \\
\hline
SimNet \cite{SIGMETRICS22:simnet, SC22:simnet} & \Uarche-dependent instruction trace & Program & Hours--days & 0.1M--600M IPS & \cmark & \xmark \\
\hline
PerfVec & \Uarche-independent instruction trace & Program & Hours & $< 1$ s & \cmark & \cmark \\
\hline
\end{tabular}
\end{table*}

Table \ref{tbl:comparison} summarizes existing studies relevant to \pve.
Ithemal \cite{ICML19:ithemal} is the first work that predicts performance using
instruction sequences, and GRANITE \cite{IISWC22:GRANITE} extends it using
graph neural network (GNN) and multiple architecture decoders.
Because modern ML models struggle with long input sequences as depicted in
Section \ref{sect:rep:challenge}, they can only deal with basic blocks with
a handful of instructions.
Taking only textual traces also makes them not suitable to predict performance
in real systems with complex memory behavior.
In performance embedding \cite{ICS23:Trumper}, GNN learns from static control
and data flow graphs as well as performance counter values to capture dynamic
execution information.
Because performance counter values are \uarch dependent, performance embedding
is not \gbl across \uarchse.
Besides, it only targets loop nests, partially because modern GNNs cannot
effectively address large graphs yet.

Program-specific models capture program characteristics inherently after
training and thus do not require any program-related input features
\cite{ASPLOS06:Ipek, HPCA07:Lee, DAC16:Li, TIST14:Chen}.
However, they come at the cost that numerous runs/simulations are required for
obtaining sufficient training data whenever encountering a new program,
hindering their generalizability.
Transferable models \cite{MICRO07:Dubach, PACT07:Khan, SC17:Marathe} alleviate
this problem by building the model of a new program from those of existing
programs, which reduce the required training data volume, but the limited
generality issue persists.

SimNet is a recent approach that accelerates simulation using ML
\cite{SIGMETRICS22:simnet, SC22:simnet}.
SimNet and \pv can both handle long instruction execution traces, but with two
crucial distinctions.
First, SimNet requires \uarche-dependent instruction features, such as cache
hit/miss, making it not \gbl across \uarchse.
In comparison, the use of \uarche-independent features makes \pv more general.
Moreover, SimNet simulates all instructions one by one in their execution order
to predict performance, while \pv only requires a simple dot product of program
and \uarch \repse, which makes it much faster.


We compare two kinds of overhead across different approaches.
Training overhead is incurred when an approach is applied to a new scenario
(e.g., new programs/\uarchse), and prediction overhead indicates how fast to
make a prediction.
Ithemal and GRANITE incur the lowest training overhead because they are limited
to basic blocks and smaller training datasets are sufficient.
Compared with other program level modeling approaches, \pv incurs less training
overhead thanks to the wide applicability of its foundation model.
Moreover, \pve's prediction speed does not depend on the program size and is
lightening fast with pre-learned \repse, similar to the speeds of
program-specific and transferable models.
To demonstrate the low overhead advantage of \pv in a quantitative way, Section
\ref{sect:case:dse} will compare \pve's speed against those of other methods in
a case of design space exploration.


Last but not least, \pv is the only one that achieves generality across both
programs and \uarchs among all methods, because it separates their performance
impact in the hierarchical model design and adopts \uarche-independent
features.


\ifdefined\showvis

\subsecspace
\subsection{Representation Visualization}
\label{sect:eval:vis}
\subsecspace

It is difficult to directly explain the meaning of learned \reps due to the
black box nature of deep neural networks.
However, indirect evidence can be found to verify the meaningfulness of \reps
produced by \pve.
We visualize the learned \reps of particular \uarchs and programs for this
purpose, by projecting high-dimensional \reps into low-dimensional spaces.

\begin{figure}[t]
\begin{minipage}{.48\linewidth}
  \centering
  \includegraphics[width=0.95\linewidth]{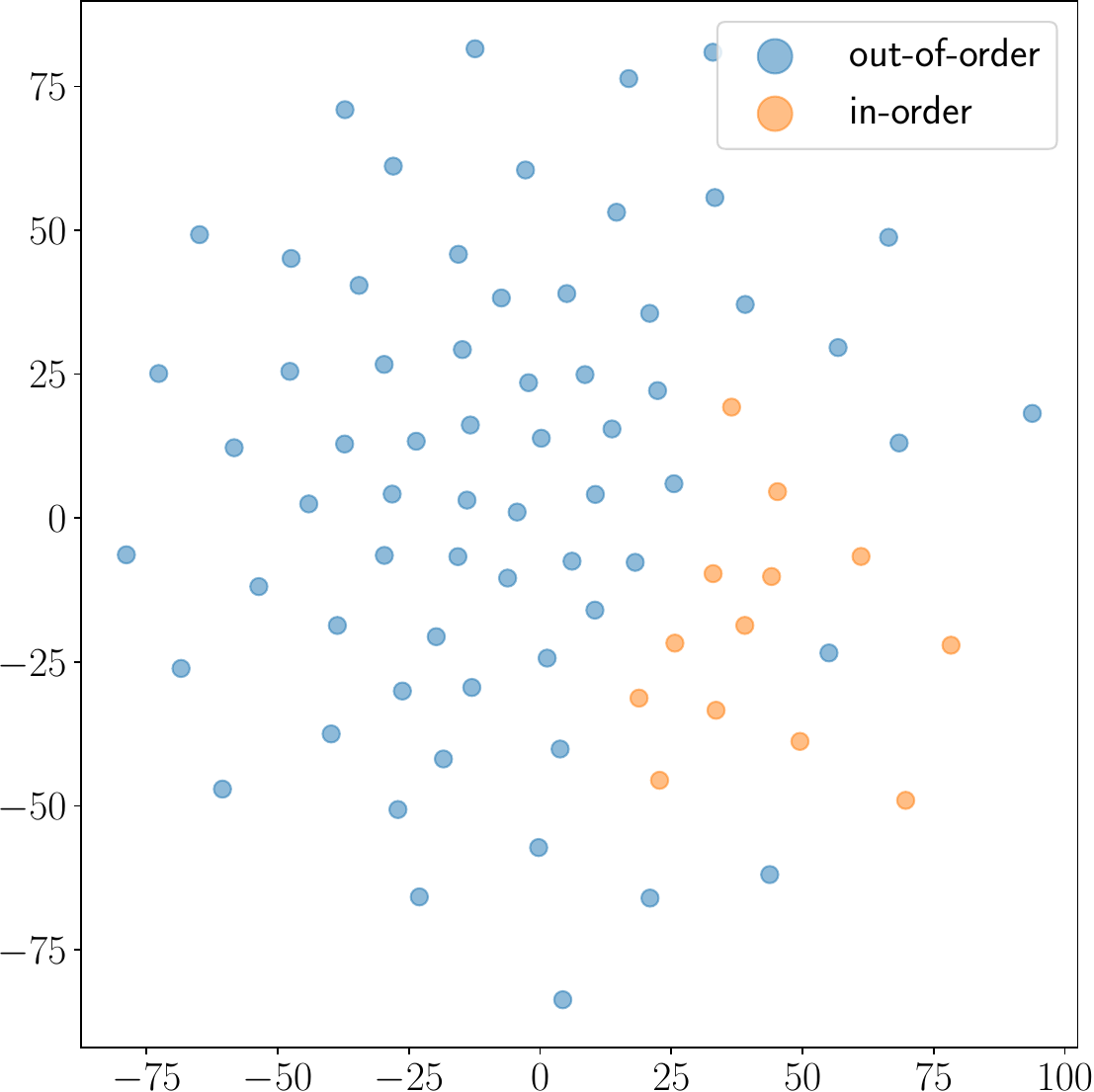}
  \caption{Two-dimensional t-SNE projection of learned \uarch \reps of various
  processors.}
  \label{fig:uarch_vis}
\end{minipage}%
\hfil
\begin{minipage}{.48\linewidth}
  \centering
  \includegraphics[width=0.95\linewidth]{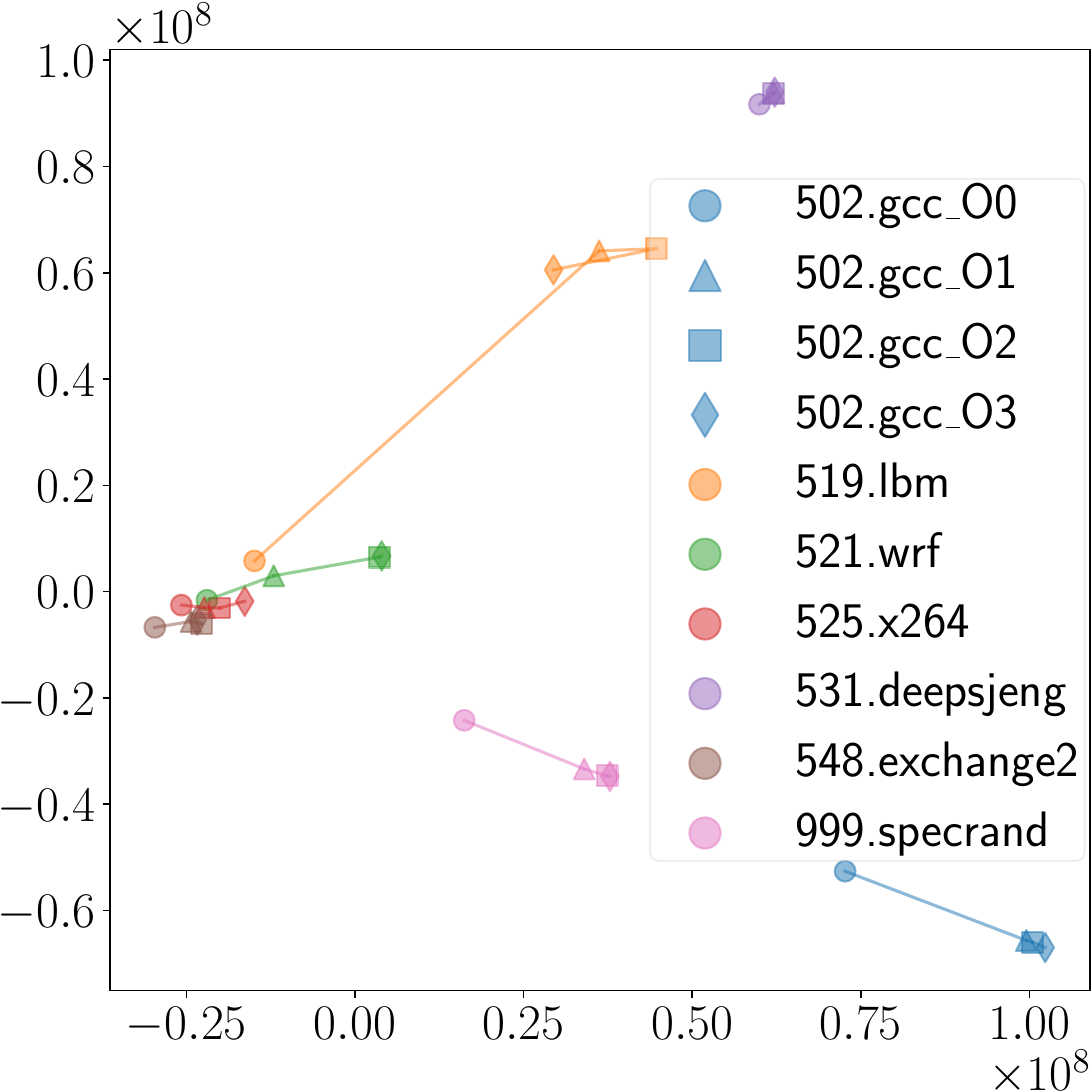}
  \caption{Two-dimensional PCA projection of learned program \reps under
  various compiler optimization levels.}
  \label{fig:opt_rep}
\end{minipage}%
\end{figure}

\begin{figure}[t]
\end{figure}

\subhead{\Uarch Representations.}
Figure \ref{fig:uarch_vis} shows the t-distributed stochastic neighbor
embedding (t-SNE) projection of 77 \uarchs \reps used in the training dataset.
Orange and blue dots denote the \reps of 13 in-order processors and 64
out-of-order processors, respectively.
Because in-order processors work quite differently compared to out-of-order
processors, we expect their \reps are somehow separate from each other.
Indeed, we observe the \reps of in-order processors are clustered
together, which makes sense due to their similarity, and are away from those of
out-of-order processors.
This indirectly proves that \pv generates meaningful \uarch \repse.


\subhead{Program Representations.}
While it is relatively easy to judge whether two \uarchs should have similar
\reps based on their configurations, it is difficult to claim so for two SPEC
benchmarks due to their complexity.
To demonstrate the foundation model produces meaningful program \repse, we
design an experiment to visualize the \reps of different versions of the same
program.
Expressly, we visualize the \reps of a program under different compiler
optimization levels and the distance between them.
This is analogous to visualizing the distance between the embeddings of related
words, e.g., countries and their capitals \cite{NIPS13:W2V}.
Because the distance in the t-SNE projection is scaled, we adopt the principal
component analysis (PCA) to project these program \repse.

Figure \ref{fig:opt_rep} shows the projected representations of different SPEC
programs under four optimization levels of gcc 8.3.1.
Cycle, triangle, square, and rhombus markers denote programs compiled using O0,
O1, O2, and O3, respectively.
Generally, the distance between two representations reflects how similar they
are in terms of performance characteristics.
Figure \ref{fig:opt_rep} includes four seen programs and three unseen programs.
The \reps of other programs are clustered around (-0.2, 0), and we omit them
for clarity.
Nevertheless, the observations below also hold for omitted programs.

There are several interesting observations.
First, when using higher optimization levels, the \rep generally moves to the
right on the x axis and to the larger absolute value direction on the y axis
(i.e., up when $y>0$ and down when $y<0$).
This trend suggests how compiler optimizations transform the \rep of a
program in general.
\texttt{519.lbm} is an outlier, and its O3 representation is on the left of
that of O2.
This is due to the challenge to learn an accurate \rep of \texttt{519.lbm} (see
Figure \ref{fig:unseen_uarch}, where it has the largest error among all
programs).
Second, the distance between O0 and O1 typically is larger than that between
other adjacent optimization levels (i.e., O1---O2, O2---O3).
This makes sense because O1 enables essential optimizations that eliminate most
unnecessary memory accesses and computations, so there is often a significant
performance boost compared with unoptimized programs (i.e., O0).
On the other hand, additional optimizations enabled by O3 are considered to
have limited effects, so we observe the \reps of O2 and O3 are quite similar
for most programs.
Last, the distance between adjacent optimization levels differs across
programs.
This reflects the fact that certain compiler optimizations help some programs
more than others.
For example, \texttt{531.deepsjeng} has a short distance between O0 and O1,
which corresponds to an 8.8\% execution time reduction under the ARM Cortex-A7
\uarche, while O1 reduces the execution time of \texttt{999.specrand} by 21.3\%
on the same \uarche, which has a longer O0---O1 distance.

These results demonstrate the meaningfulness of learned program and \uarch
\repse.
As a result, besides using program and \uarch \reps together for performance
predictions, they can also be used for broad downstream tasks that require
quantitative and/or qualitative performance analysis.

\fi

\secspace
\section{Applications}
\label{sect:case}
\secspace

\pv can be applied in many domains where performance modeling plays a critical
role.
We will illustrate a design space exploration (DSE) case and a program
analysis case.

\subsecspace
\subsection{Design Space Exploration}
\label{sect:case:dse}
\subsecspace




\subhead{DSE Workflow.}
Leveraging the pre-trained foundation model, we can quickly explore any \uarch
design space of interest by training a corresponding \uarch \rep model.
The proposed \uarch DSE procedure works by \circled{1} sampling the \uarch
design space and simulating several programs to obtain instruction latencies
under sampled \uarchse, which serves as the training dataset.
Notably, short simulations can generate a sufficient training set that includes
many instructions, and the programs used for training are not necessarily the
target programs due to the generality of \pve.
\circled{2} Using the data obtained in \circled{1}, we train a \uarch \rep
model to take \uarch parameters as inputs and output the corresponding \uarch
\repse.
The framework to train the \uarch \rep model is similar to the one depicted in
Figure \ref{fig:instrep} with the instruction \rep model frozen.
The difference is we replace the \uarch \rep table with a \uarch \rep model
that generates \reps from input parameters, so that it can generalize to unseen
\uarchse.
The training is fast because the pre-trained foundation model does not get
updated.
\circled{3} The trained \uarch \rep model can generate the \reps of all
possible \uarch configurations, which are in turn used to predict the
performance of any \uarch and program pair with a simple dot product.
The ``optimal'' design can be selected using prediction results.

\subhead{L1 and L2 Cache Size DSE.}
As a case study, we explore the design space of L1 data and L2 cache sizes for
17 SPEC programs while using the ARM Cortex-A7 model as the core and fixing
other configurations.
The L1 data cache size varies from 4~kB to 128~kB, and the L2 cache size varies
from 256~kB to 8~MB.
Both sizes are constraint to be powers of two.
Together, there are $6 \times 6 = 36$ configurations.

We assume the objective function to minimize is $(1000 + 10 \times
\text{L1\_size\_in\_kB} + \text{L2\_size\_in\_kB}) \times
\text{execution\_time}$.
This objective can be interpreted as finding the optimal cache capacities that
minimize the total chip footprint without significant performance loss, for
every program.
Artificial constants and coefficients are used for illustration purposes
without losing generality: A multiplier of 10 is applied to the L1 cache size
due to the bigger L1 SRAM cells, and the constant 1000 represents the area
overhead of other on-chip components.

We adopt a simple 2-layer MLP with 4.4~k parameters as the \uarch \rep model,
and its inputs are L1 and L2 cache sizes.
We simulate three SPEC programs by 100 million instructions each on 18 sampled
\uarchs to obtain its training dataset.

\subhead{Results.}
Out of all 17 target programs, the cache sizes predicted to minimize the
objective function by \pv are the optimal ones for four of them, among the top
two choices for 11 of them, the top three choices for 15 of them, and the top
five choices for all.
On average, only 3.6\% designs outperform the design selected by \pve,
demonstrating its accuracy.

\begin{figure}[t]
\centering
    \begin{subfigure}[t]{0.49\linewidth}
        \centering
        \includegraphics[width=\linewidth]{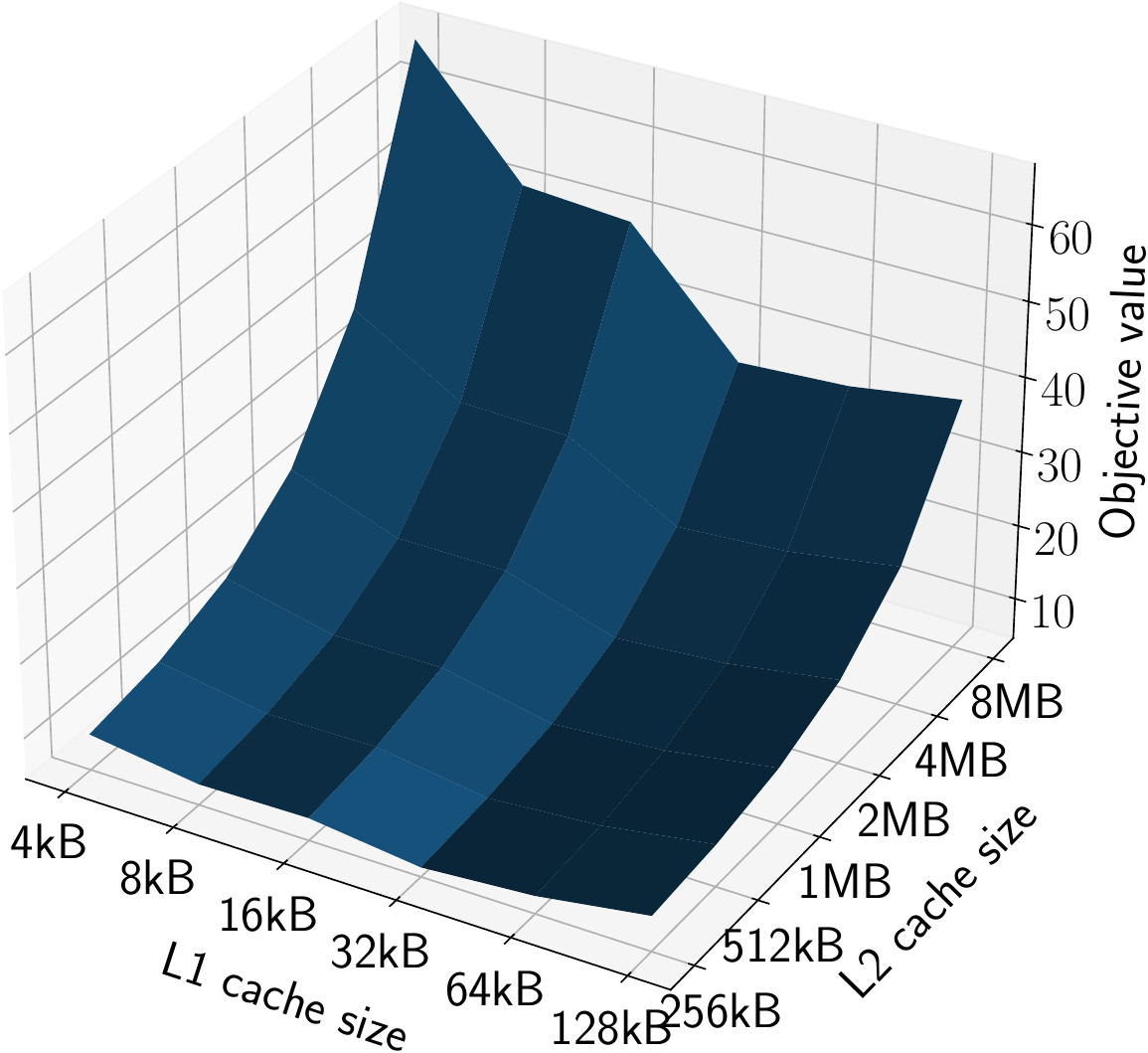}
        \caption{gem5.}
    \end{subfigure}%
    \hfil
    \begin{subfigure}[t]{0.49\linewidth}
        \centering
        \includegraphics[width=\linewidth]{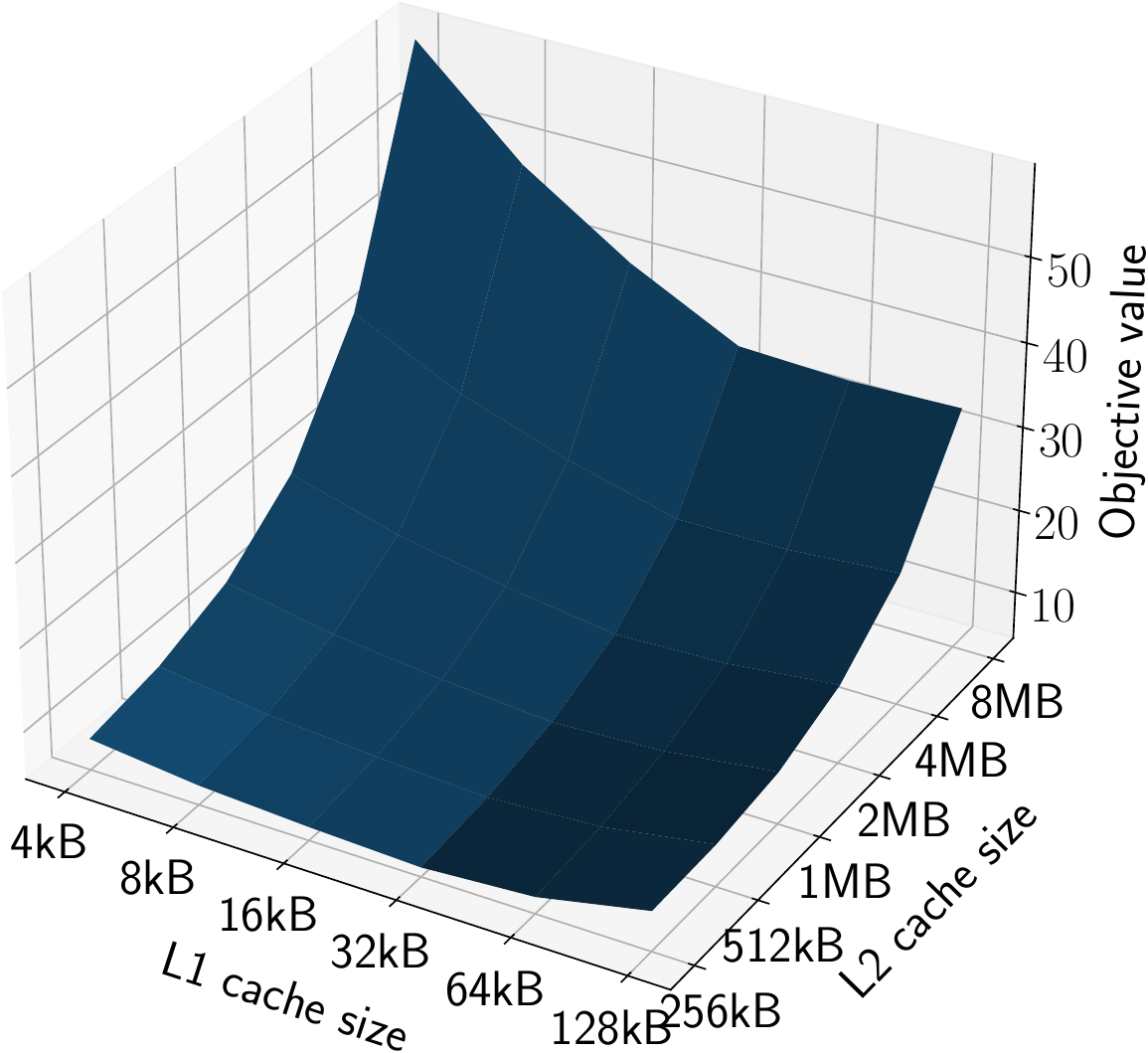}
        \caption{\pve.}
    \end{subfigure}
\caption{Objective function values of \texttt{508.namd} across the cache size search space.}
\label{fig:cache_dse}
\end{figure}

Figure \ref{fig:cache_dse} shows the predicted objective function values under
various L1 and L2 cache sizes and compares it with those obtained using
exhaustive gem5 simulation for \texttt{508.namd} as an example.
While the objective function surfaces of \pv and gem5 have similar shapes, an
intersting observation is the predicted surface of \pv is smoother than that of
gem5.
This stems from the use of a simple 2-layer MLP \uarch \rep model, which tends
to introduce less nonlinearity to prediction results.
Nevertheless, \pv correctly projects the trajectories and that a 32 kB L1 data
cache works best for all L2 sizes, and large L2 caches do not benefit this
benchmark.

\pve's overhead comes from the data collection and the \uarch \rep model
training, and its prediction time is negligible.
In total, \pv takes 11 hours to accomplish this cache size DSE, including five
hours of gem5 simulation for data collection and six hours to train the \uarch
\rep model.

\subhead{Comparison with State-of-the-Art DSE Approaches.}
Simulation is the most widely used and default method in \uarch DSE.
Because the time overhead of simulation increases linearly with the number of
target programs $\times$ the number of \uarch configurations, it is impractical
to use simulation for large-scale DSE.
In this experiment, to simulate each of 17 programs by one billion instructions
on 36 cache configurations, gem5 requires roughly 600 hours, which is about $50
\times$ slower than \pve.
Simulating these programs to the end will require tens of thousands of hours.

SimNet is a recent ML-based simulation acceleration approach
\cite{SIGMETRICS22:simnet, SC22:simnet}.
Its simulation involves two steps.
First, gem5 simulation with simplified processor and cache models is performed
to gather SimNet's input traces.
This is necessary because unlike \pve, SimNet's input trace requires
\uarche-dependent information (e.g., cache hit/miss).
This process takes roughly 89 hours in this DSE case.
Second, SimNet simulates all traces using a trained ML model, which takes about
81 hours in this case using the same GPU machine.
Together, SimNet requires 170 hours for this DSE, about $15 \times$ slower than
\pve.
Moreover, this is an optimistic cost estimation of SimNet because its training
overhead is not considered, which may take days \cite{SIGMETRICS22:simnet}.

\begin{table*}[t]
\footnotesize
\centering
\caption{Comparison of ML-based DSE methods. Overhead includes both model
construction and prediction time (hours), and quality is measured by how close
the selected design is to the optimal design (smaller is better).}
\label{tbl:dse}
\begin{tabular}{|c|c|c|c|c|}
\hline
 & MLP predictor \cite{ASPLOS06:Ipek} & Cross-program predictor \cite{MICRO07:Dubach} & ActBoost \cite{DAC16:Li} & PerfVec \\
\hline
\hline
Overhead & 150 & 84 & 170 & 11 \\
\hline
Quality & 4.4\% & 4.7\% & 3.6\% & 3.6\% \\
\hline
\end{tabular}
\end{table*}

To reduce the simulation workload, previous research has proposed to train
ML-based predictive models using selective simulation results.
Table \ref{tbl:dse} compares \pv with representative ones, including MLP
predictors \cite{ASPLOS06:Ipek}, cross-program linear predictors
\cite{MICRO07:Dubach}, and MLP-based AdaBoost models \cite{DAC16:Li}.
These approaches need to simulate every target programs on a significant amount
of \uarch configurations for model training.
To achieve a comparable quality of \pve, they require simulating at least 25\%,
14\%, and 28\% of the design space, respectively.
These simulations take 84--170 hours, which are 8--15$\times$ of \pve's cost.

While the overhead of all above approaches increases with the amount and sizes
of target programs, \pv has a constant overhead with respect to the program
count and sizes thanks to the reusability of program \repse.
As a result, the speed advantage of \pv will be more significant when
evaluating more and larger programs.

\begin{figure}[t]
\centering
\includegraphics[width=0.8\linewidth]{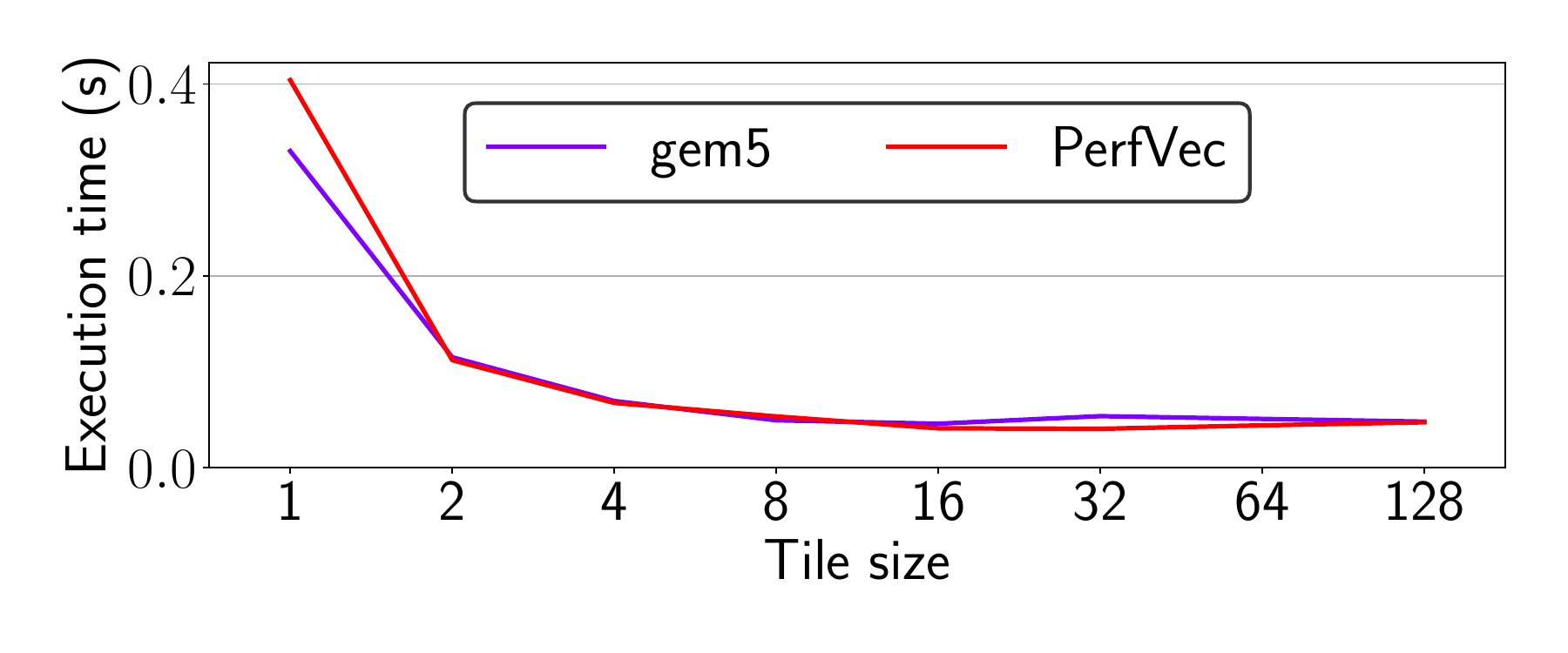}
\caption{MM performance under various tile sizes.}
\label{fig:mmtile}
\end{figure}

\subsecspace
\subsection{Loop Tiling Analysis}
\label{sect:case:tile}
\subsecspace

The \pv framework also can be used for program analysis and optimization.
As an example, we devise an experiment to analyze the effect of loop tiling for
matrix multiplication (MM).
A naive implementation of MM has three nested loops to calculate and accumulate
the product of source matrix elements.
Our loop tiling implementation blocks all three loops, and a uniform tile size
is adopted for simplicity.

Figure \ref{fig:mmtile} compares the execution time of MM generated by gem5 and
\pv with tile sizes from 1 to 128.
The gem5 Cortex-A7 configuration is used as the processor model.
The matrix size is $256 \times 256$.
Overall, the prediction results of \pv agree with those of the gem5 simulation.
With larger tile sizes, wider vector instructions can be leveraged for better
performance, resulting in sharp execution time decrements until the tile size of
8.
When a tile exceeds the L1 data cache size, performance degradation is observed
due to increasing L1 cache misses, which is less significant in this case
because all matrices fit into the L2 cache.
The optimal tile size is 16 under gem5, while \pv predicts tile sizes of 16 and
32 having similar and the best performance.
Of note, this analysis incurs negligible inference overhead and no training
overhead because the pre-trained foundation model is used.

\ifdefined\showphase

\begin{figure*}[t]
\centering
%
\begin{minipage}[t]{\textwidth}
\centerline{
  \includegraphics[width=0.25\textwidth]{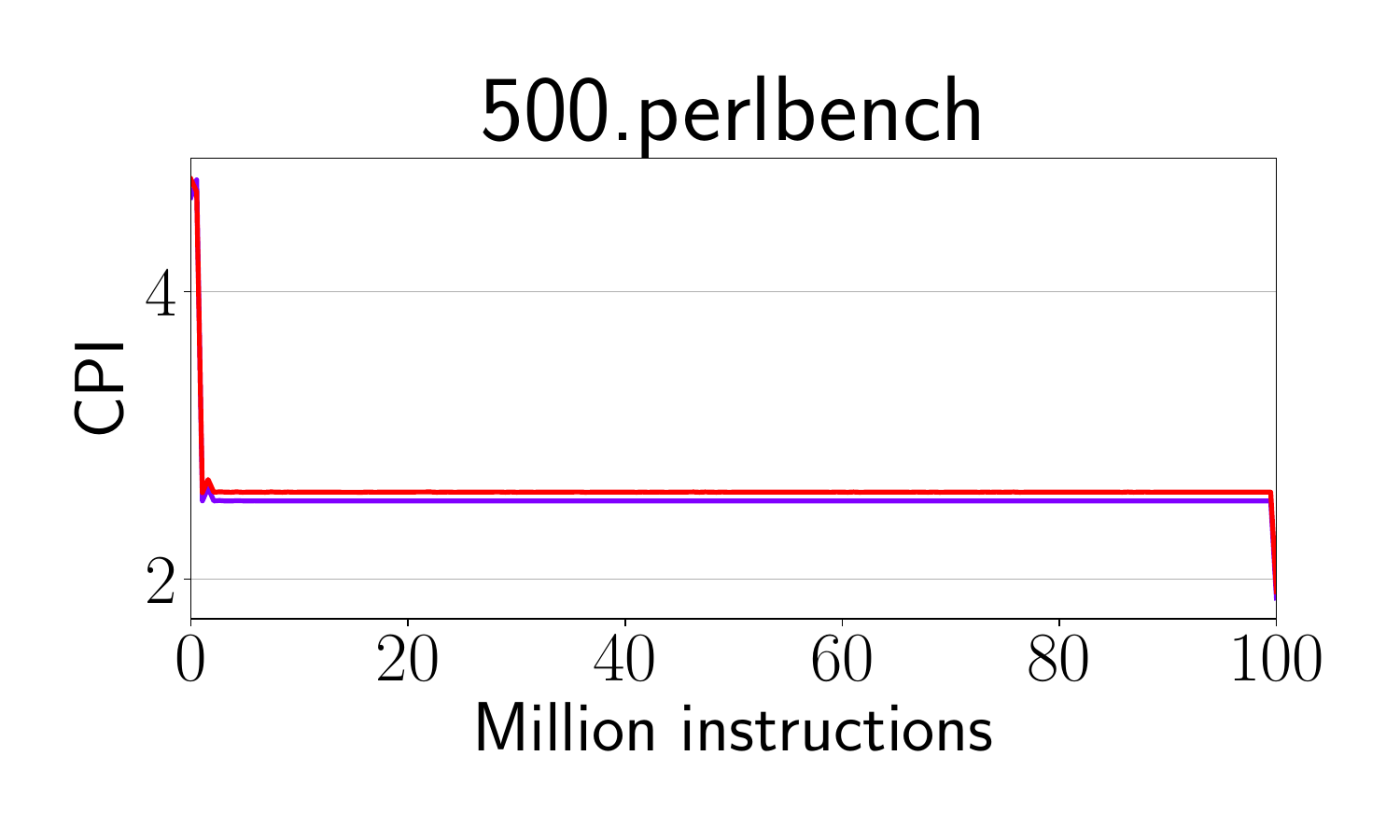}
  \hfil
  \includegraphics[width=0.25\textwidth]{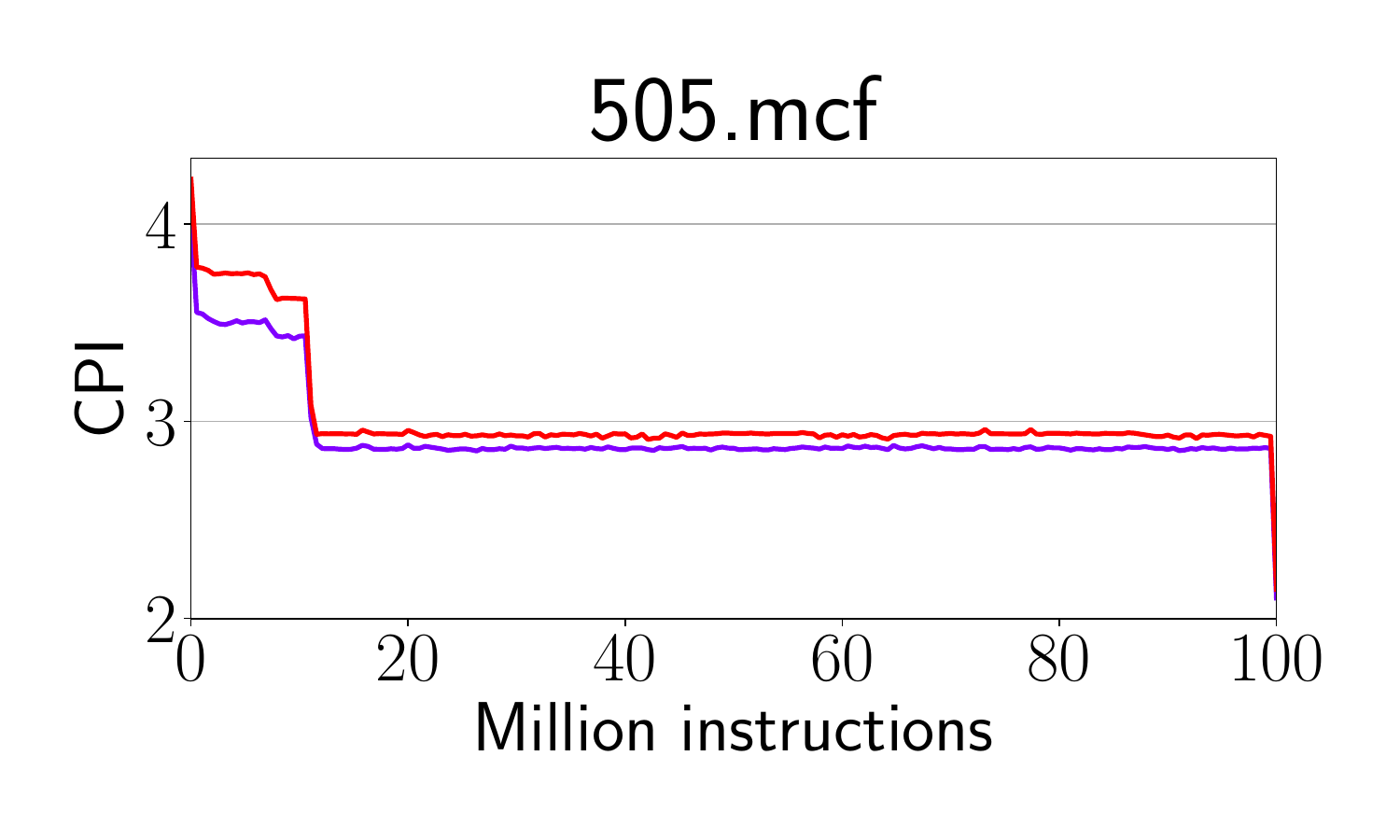}
  \hfil
  \includegraphics[width=0.25\textwidth]{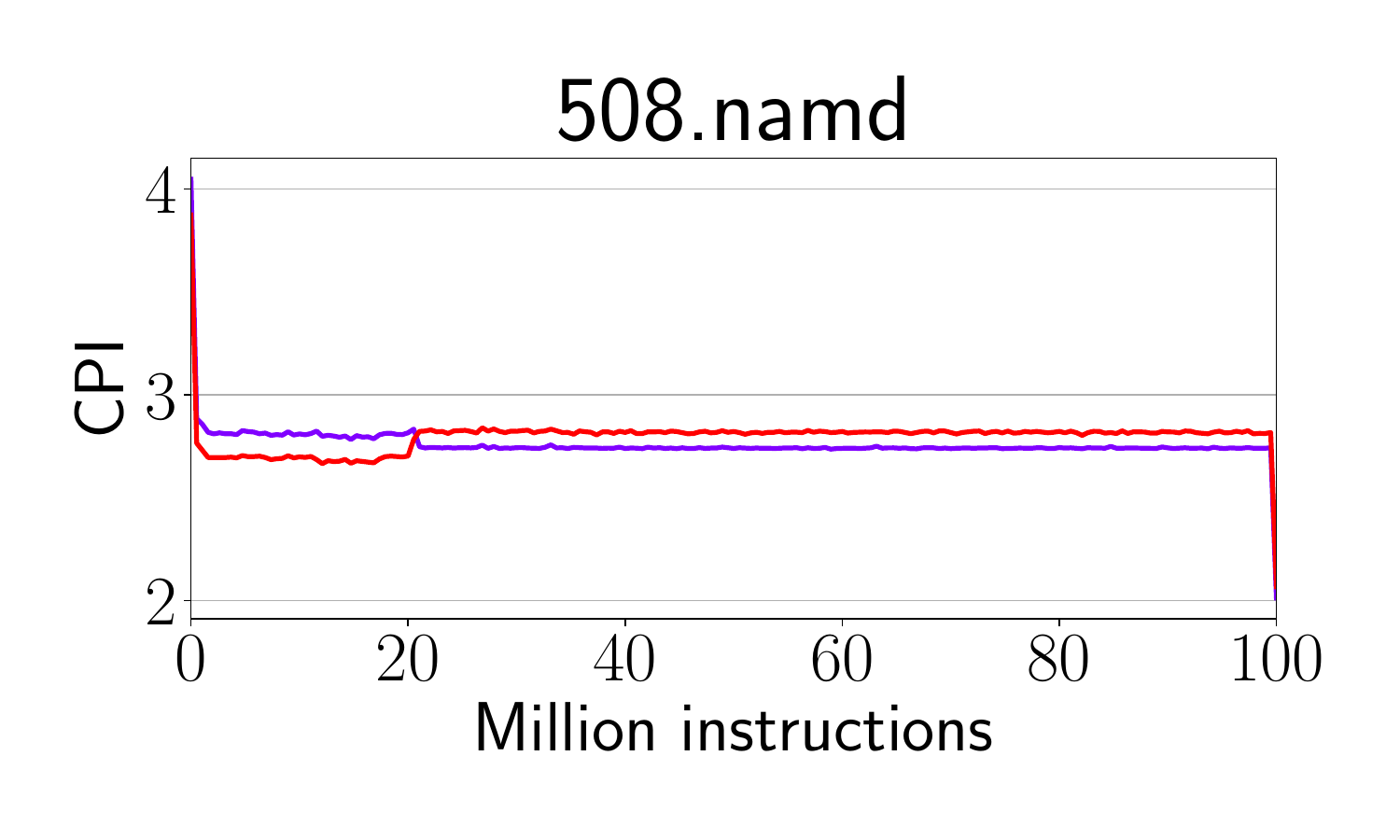}
  \hfil
  \includegraphics[width=0.25\textwidth]{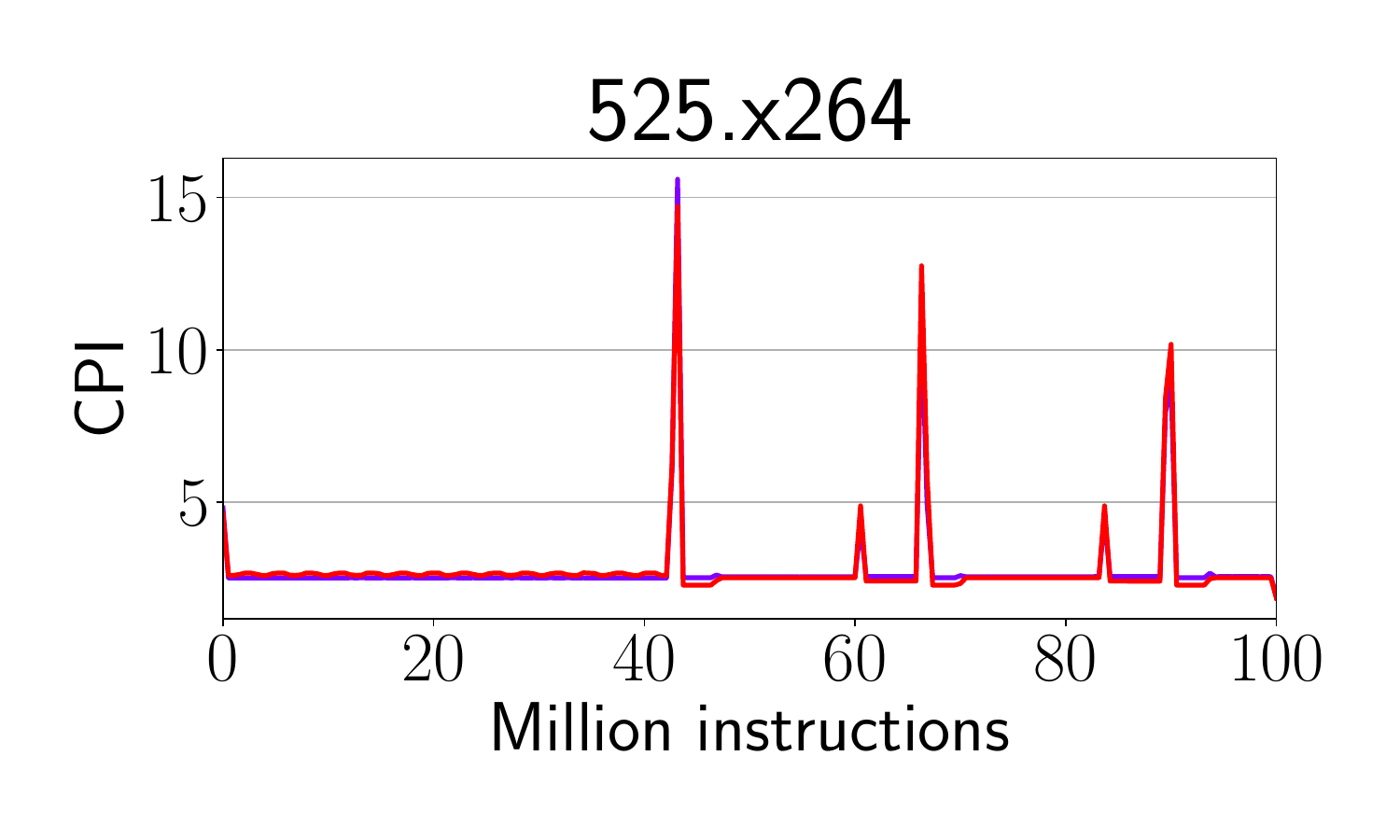}
}
\end{minipage}%

\begin{minipage}[t]{\textwidth}
\centerline{
  \includegraphics[width=0.25\textwidth]{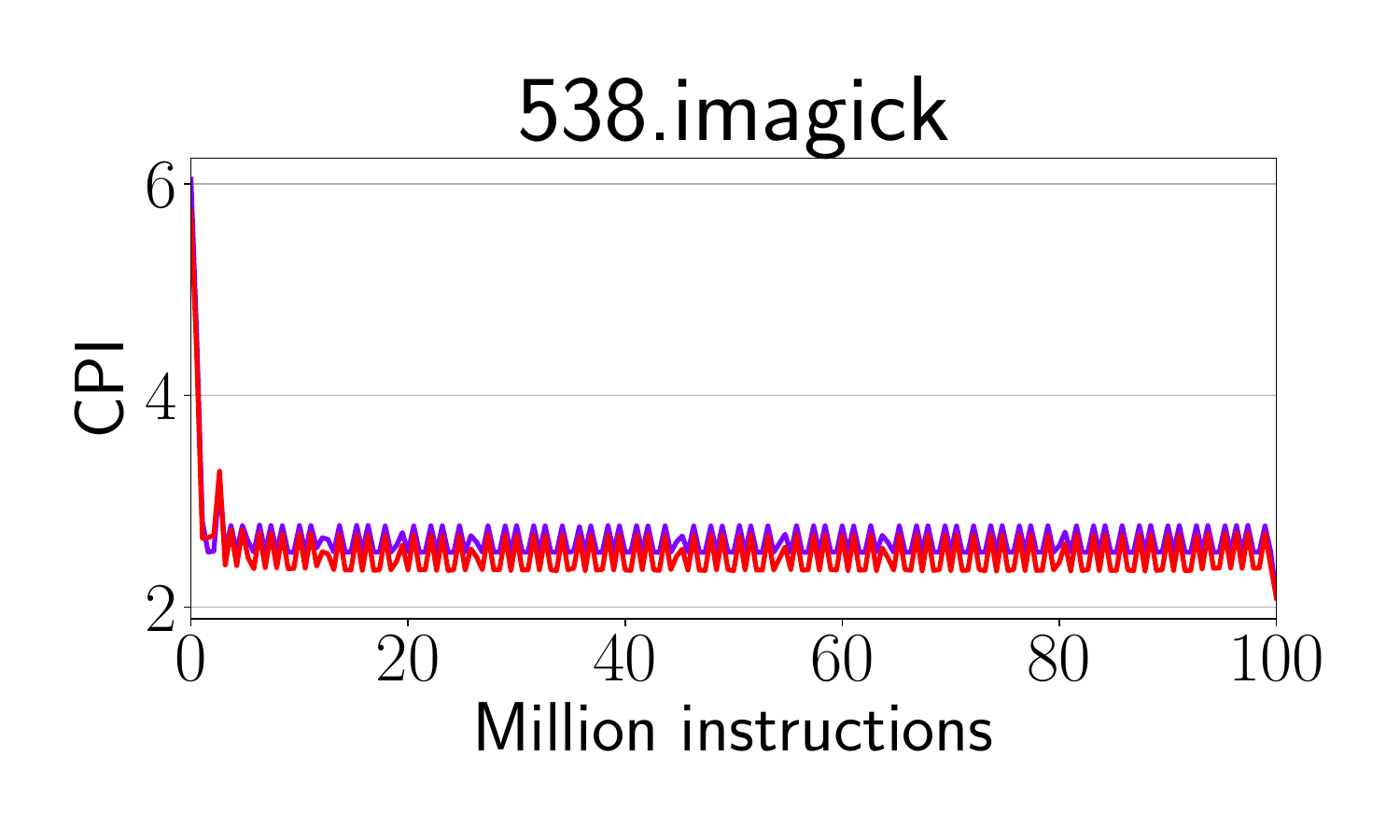}
  \hfil
  \includegraphics[width=0.25\textwidth]{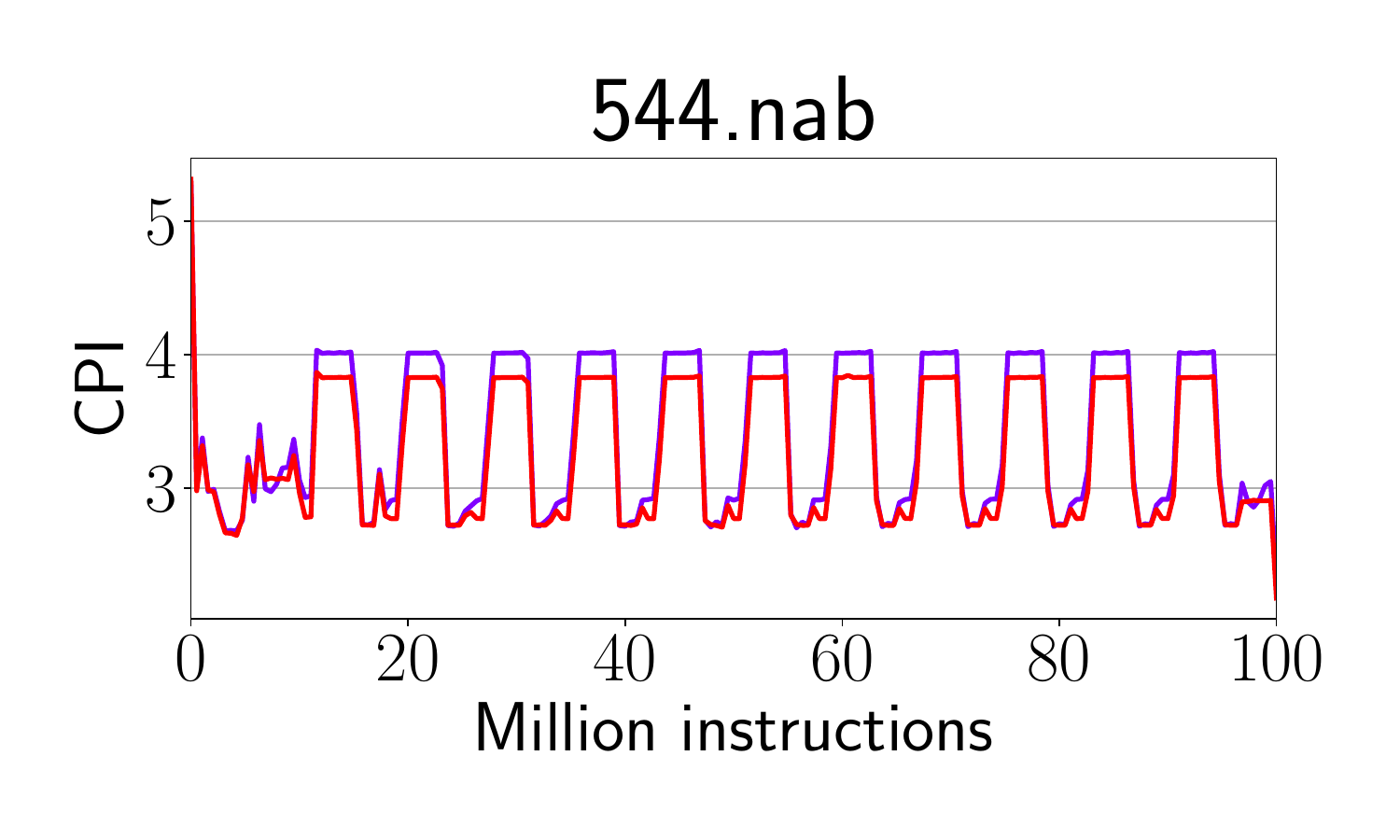}
  \hfil
  \includegraphics[width=0.25\textwidth]{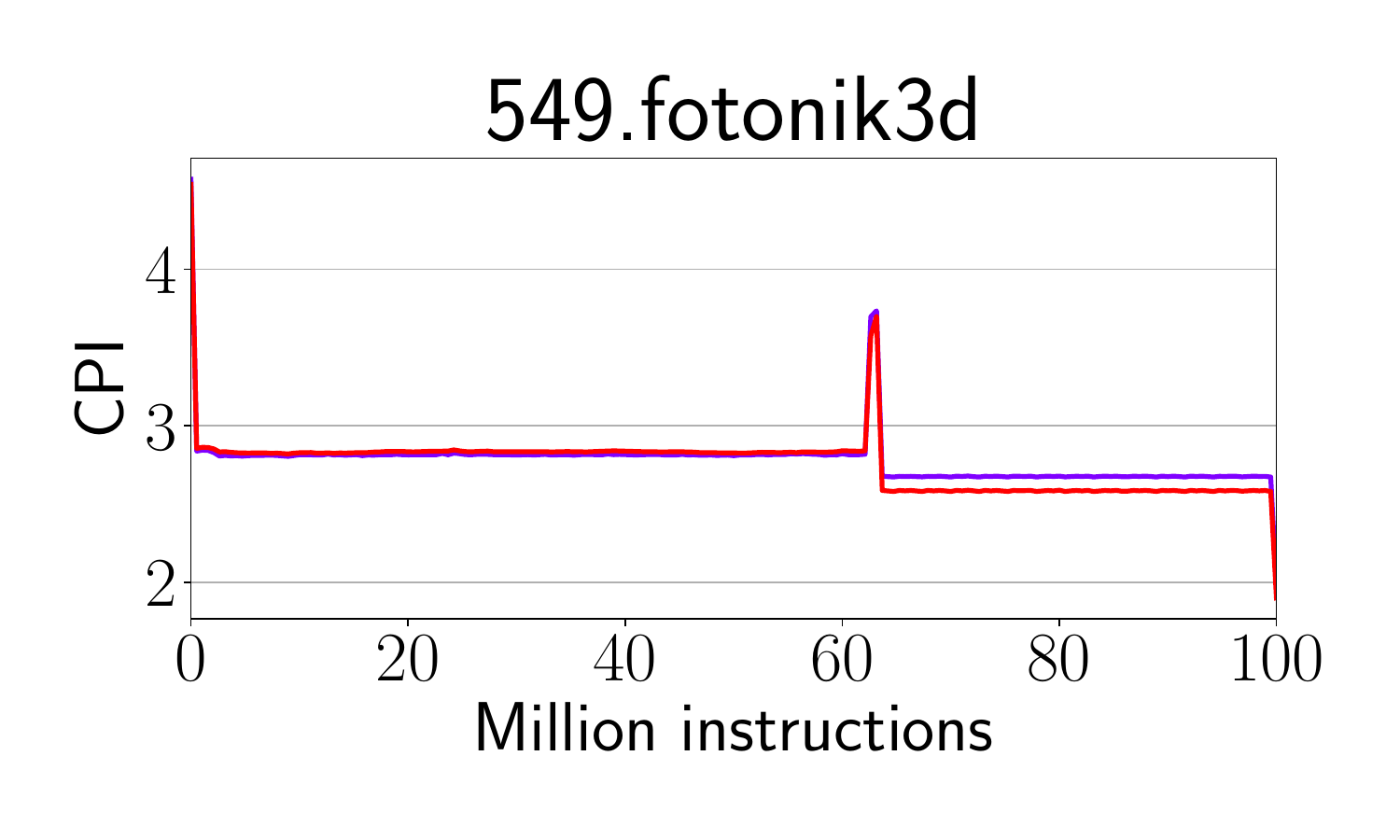}
  \hfil
  \includegraphics[width=0.25\textwidth]{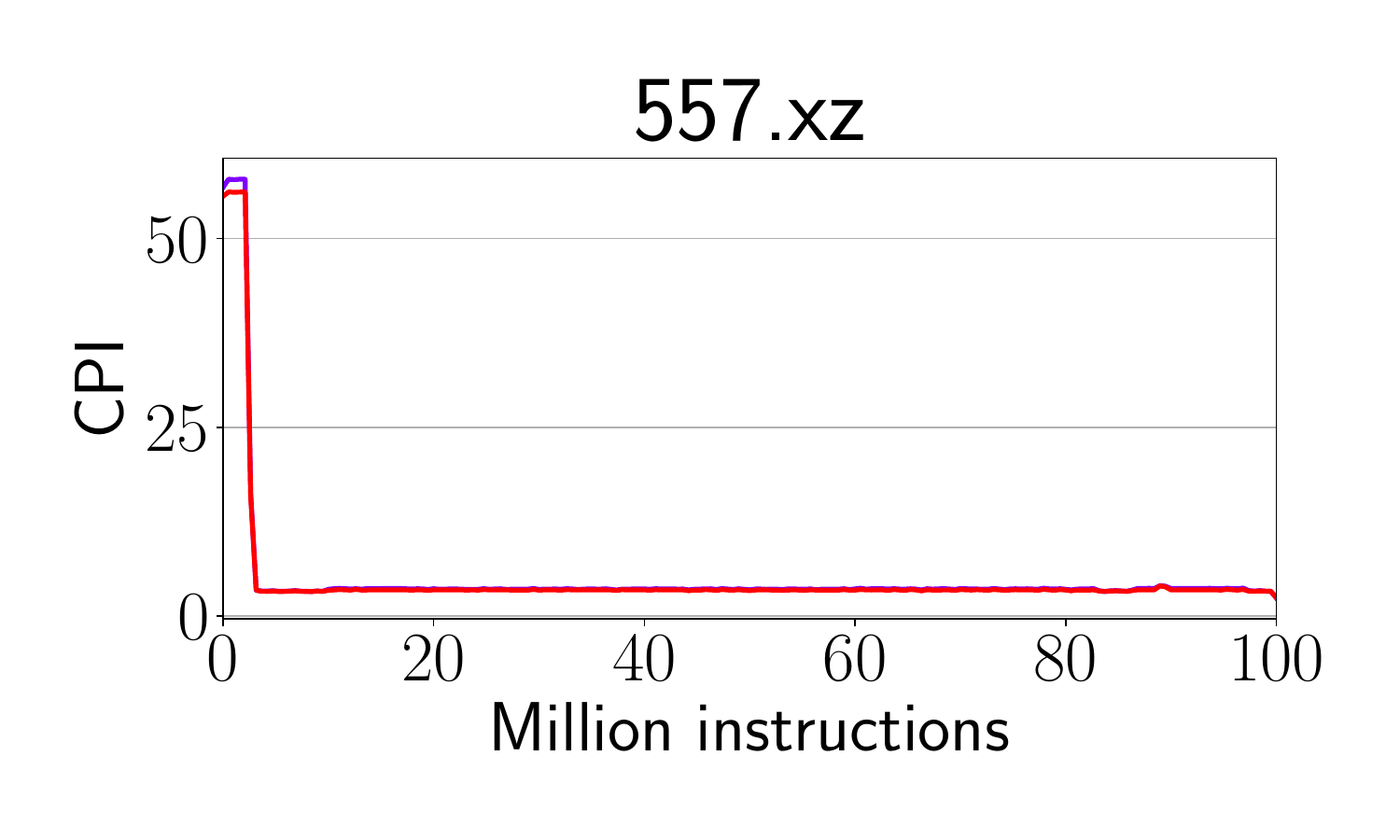}
}
\end{minipage}%

\caption{CPI curves generated by gem5 simulation (purple) and \pv (red) over
100 million instructions of SPEC programs.}
\label{fig:cpicurve}
\end{figure*}

\subsecspace
\subsection{Phase-level Performance Analysis}
\label{sect:case:phase}
\subsecspace

Because of the compositional property of instruction \repse, we not only can
learn the \rep of a whole program, but also extract execution phase \repse.
A phase \rep is constructed in a similar way as that of program \repse, by
summing the \reps of all instructions with it.
To illustrate this use case, we partition the execution of 100 million
instructions into segments of 0.5 million instructions.
Then, we extract the \reps of these segments to predict the cycle per
instruction (CPI) variation during execution.

Figure \ref{fig:cpicurve} shows the CPI curves calculated using phase-level
\reps and compares them to those obtained using gem5 simulation for eight
programs.
A randomly generated \uarch configuration in the training dataset is used here.
The predicted CPI curves of \pv capture the overall performance variations well
for all programs, and they thus can help identify potential performance
  bottlenecks.
This observation also holds for other \uarchs and programs that are not shown
in Figure \ref{fig:cpicurve} due to the space constraint.
Among other performance modeling approaches, only expensive simulation can
provide such execution details, and existing data-driven modeling approaches
cannot accomplish it easily because they are not compositional.


\fi

\secspace
\section{Related Work}
\label{sect:relatedwork}
\secspace



\subhead{ML-based Performance Modeling.}
Ipek, Lee, \etal are the pioneers who constructed neural network and regression
models to predict program-level performance under a broad range of \uarch
configurations \cite{ASPLOS06:Ipek, ASPLOS06:Lee, HPCA07:Lee, PPoPP07:Lee}.
COAL combines semi-supervised and active learning to train models
\cite{TIST14:Chen}.
ActBoost integrates statistical sampling and active AdaBoost learning to reduce
training overhead \cite{DAC16:Li}.
Once-for-all trains neural nets to predict the latency and accuracy of a set of
neural architectures \cite{ICLR19:OFA}.
Hutter and Solomonik use tensor decomposition for performance modeling
\cite{SC23:Hutter}.
Wu \etal use performance counters to predict GPU performance and power
\cite{HPCA15:Wu}.
\cite{TACO18:Wang} predicts the performance of certain benchmark suites for
Intel CPUs.
There also has been research on predicting processor performance/power based on
those obtained on different types of processors \cite{MICRO15:Ardalani,
SBAC-PAD14:Baldini, TECS17:Oneal} or with varied ISAs \cite{zheng2015learning,
DAC16:Zheng}.
To lower the cost of training data acquisition, transfer learning leverages
existing models on other programs \cite{MICRO07:Dubach, PACT07:Khan} or small
scale runs \cite{SC17:Marathe}.

Ithemal \cite{ICML19:ithemal} uses LSTM to predict the execution latency of
static basic blocks composed by a few instructions assuming ideal caches.
Using it as a surrogate model, DiffTune \cite{MICRO20:DiffTune} searches
simulator parameters for target \uarchs through gradient decent.
GRANITE \cite{IISWC22:GRANITE} models basic block performance using GNN and
architecture dependent decoders.
These methods are limited to basic blocks with a handful of instructions rather
than lengthy programs.
They also do not consider dynamic execution behaviors such as memory accesses
and branches.
Tr\"{u}mper \etal propose to learn the performance embeddings of parallelizable
loop nests which are in turn used to guide compiler optimizations
\cite{ICS23:Trumper}.
SimNet \cite{SIGMETRICS22:simnet, SC22:simnet} predicts individual instruction
latencies given execution context and then predicts program performance by
simulating all executed instructions.

In existing ML-based modeling and simulation approaches, trained models are
bounded to certain programs and/or \uarchse, which limits their generality.
In comparison, \pv is the first that achieves generality across both programs
and \uarchse.



\subhead{Analytical Performance Modeling.}
Roofline models the relationship between computation and memory throughputs
\cite{CACM09:Roofline}.
\cite{TOCS09:Eyerman} builds an interval-based model to estimate the
impact of caches and branches.
\cite{TC16:Steen} uses \uarche-independent program features, and \pv adopts
some of them.
\cite{ISPASS11:Eyerman} uses backward regression to infer unknown parameters of
analytical models.
The model in \cite{SC02:Snavely} convolves application and machine signatures
to predict performance.
\cite{Computer09:Barker} builds program-specific models to predict the
performance of large-scale supercomputers.
Palm semi-automatically generates performance models through source code
annotations \cite{ICS14:Palm}.
Due to the extreme complexity of modern processors, these human-designed
analytical models cannot capture all performance-related execution details and
therefore often fail to be accurate.
Instead, \pv extracts and learns these details from abundant instruction
execution traces to minimize accuracy loss.


\subhead{Learning Semantic Code Representations.}
Recent research has applied ML techniques developed for NLP to understand the
semantics of program code \cite{CSUR19:Allamanis}.
CuBERT learns token \reps using surrounding tokens as
contextual inputs \cite{ICML20:Kanade}, and CodeBERT takes both code and their
natural language descriptions into account \cite{EMNLP20:CodeBERT}.
\cite{ICML15:Piech} proposes to learn program embeddings by mapping program
states before execution to those after execution. 
\cite{ICLR18:Wang} offers to learn program embeddings using dynamic variable
valuations, and \cite{PLDI20:Wang} extends it by integrating symbolic execution
traces.
To learn LLVM intermediate representation (IR) embeddings, \cite{NIPS18:NCC}
assumes IR instructions that are close in data or control flow have similar
semantics, and IR2VEC takes into account data dependency information
\cite{TACO20:IR2VEC}.
Several recent work explores the use of GNNs \cite{ICLR18:Allamanis,
ICLR20:Shi, ICML21:ProGraML}.

Research into semantic \reps is useful for tasks such as program
classification, completion/repair, and similarity detection \cite{copilot}.
These semantic \reps are not designed to model performance.
In comparison, \pv learns \reps that capture performance implications, and in
turn enables performance-related analysis and optimization tasks such as design
space exploration.



\secspace
\section{Conclusion}
\label{sect:conclusion}
\secspace

This paper presents \pve, a novel and generic performance modeling framework
that exploits deep learning to autonomously separate the performance impact of
programs and \uarchse.
The approach enabled by \pv has general applicability across a broad spectrum
of programs and \uarchse, as well as speed advantages.
While this paper focuses on sequential program modeling, it serves as the
foundation to model parallel programs.
We plan to extend \pv for that purpose in the future.


\section*{Acknowledgments}

We would like to thank the paper and artifact reviewers for their helpful
feedbacks.
We would also like to thank Charity Plata for her editing assist.
Google Bard/Gemini was used to provide editing suggestions for several
sentences in the paper.
This research was conducted at the Brookhaven National Laboratory, supported by
the U.S. Department of Energy’s Office of Science under Contract No.
DE-SC0012704.
This research used resources (Wombat) of the Oak Ridge Leadership Computing
Facility at the Oak Ridge National Laboratory, which is supported by the Office
of Science of the U.S. Department of Energy under Contract No.
DE-AC05-00OR22725.

\bibliographystyle{IEEEtranS}
\bibliography{ml, arch}


\end{document}